\documentclass{article}
\usepackage{graphicx} 
\usepackage{microtype}
\usepackage{graphicx}
\usepackage{bbm}
\usepackage{subfigure}
\usepackage[table,xcdraw]{xcolor} 
\usepackage{array} 
\usepackage{booktabs} 
\usepackage{etoc} 
\usepackage{amsmath}
\usepackage{amssymb}
\usepackage{mathtools}
\usepackage{amsthm}
\usepackage{hyperref}
\usepackage[a4paper, margin=1in]{geometry}

\title{\rule{\linewidth}{1.0 pt} \\ 
       \vspace{0.5em}              
       \textbf{A Deep Learning approach for parametrized and time dependent Partial Differential Equations using Dimensionality Reduction and Neural ODEs}
       \vspace{0.1em} \\           
       \rule{\linewidth}{1.0pt}}   
\author{
    \textbf{Alessandro Longhi}\thanks{a.longhi@tudelft.nl. The source code is available \href{https://github.com/Aleartulon/AE_NODE}{here}.
}\\
    \small{Delft University of Technology}\\ \small{Department of Radiation Science and Technology}\\ \small{Delft, The Netherlands} 
    \and
    \textbf{Danny Lathouwers}\\
    \small{Delft University of Technology}\\ \small{Department of Radiation Science and Technology}\\ \small{Delft, The Netherlands} 
    \and
    \textbf{Zoltán Perkó}\\
    \small{Delft University of Technology}\\ \small{Department of Radiation Science and Technology}\\ \small{Delft, The Netherlands}} 
\date{}
\begin{document}
\maketitle
\begin{abstract}
Partial Differential Equations (PDEs) are central to science and engineering. Since solving them is computationally expensive, a lot of effort has been put into approximating their solution operator via both traditional and recently increasingly Deep Learning (DL) techniques. A conclusive methodology capable of accounting both for (continuous) time and parameter dependency in such DL models however is still lacking. In this paper, we propose an autoregressive and data-driven method using the analogy with classical numerical solvers for time-dependent, parametric and (typically) nonlinear PDEs. We present how Dimensionality Reduction (DR) can be coupled with Neural Ordinary Differential Equations (NODEs) in order to learn the solution operator of arbitrary PDEs. The idea of our work is that it is possible to map the high-fidelity (i.e., high-dimensional) PDE solution space into a reduced (low-dimensional) space, which subsequently exhibits dynamics governed by a (latent) Ordinary Differential Equation (ODE). Solving this (easier) ODE in the reduced space allows avoiding solving the PDE in the high-dimensional solution space, thus decreasing the computational burden for repeated calculations for e.g., uncertainty quantification or design optimization purposes. The main outcome of this work is the importance of exploiting DR as opposed to the recent trend of building large and complex architectures: we show that by leveraging DR we can deliver not only more accurate predictions, but also a considerably lighter and faster DL model compared to  existing methodologies.
\end{abstract}
\newpage
\section{Introduction}
Physical simulations are crucial to all areas of physics and engineering, such as fluid dynamics, nuclear physics, climate science, etc. Although a lot of work has been done in constructing robust and quick numerical Partial Differential Equations (PDE) solvers \cite{numerical_matemathics}, traditional solvers such as finite element methods are still computationally expensive when the system is complex. This is a problem especially when repeated evaluations of a model for different initial conditions and parameters are needed, which is typical in sensitivity analysis, design optimization or uncertainty quantification studies \cite{UQ,Perk2014}. To overcome such time limitations, decades of extensive research has been put into building so called \textit{surrogate models}, i.e., faster to evaluate but accurate enough approximations of the original complex model describing the physical system of interest. 



The first studies on surrogate modeling fall under the umbrella of \textit{Reduced Order Modeling} (ROM) \cite{ROM} methods, with the pioneering work on Proper Orthogonal Decomposition (POD) by Lumley in 1967 \cite{pod_lumley}. The main assumption of ROM is that a system determined by $N$ (potentially infinite) degrees of freedom (full space) can instead be projected into a lower dimensional space of dimension $n$ (reduced space), hence its evolution can be calculated by solving a much smaller system of $n\ll N$ equations. 
A common way of proceeding, under the name of reduce basis methods, is by assuming that the solution field $s(x,t)$ of a PDE can be approximated as: $s(x,t)\approx\sum_{k=1}^n a_k(t)V_k(x)$ with $a_k(t)$ being time-dependent coefficients and $V_k(x)$ being independent variable $x$ dependent functions, the latter constituting an orthonormal basis (the reduced basis). Once the optimal basis is found the system is completely described by the $n$ coefficients $a_k(t)$. 
The same concept of Dimensionality Reduction (DR) is known in the Deep Learning (DL) field under the name of \textit{manifold hypothesis} \cite{Fefferman2016, goldt2020modeling, cohen2014learning, higgins2018towards}, analogously stating that high-dimensional data typically lie in low dimensional manifolds (due to correlations, symmetries, noise in data, etc.). In DL jargon, this reduced space is usually named \textit{latent} space.

Several recent works explore the potential of DL for surrogate modeling, both following the ideas of traditional ROM approaches and proposing new paradigms. A non exhaustive list of methods that integrate DL techniques with ROM concepts is provided in \cite{FRESCA2022114181, bhattacharya2021model, romcnn, SoleraRico2024}. Among these works, DL is used to approximate the mapping between full space and reduced space, to determine the coefficients of the reduced basis and/or to map initial states of the PDE into the PDE solution $s(x,t)$. \cite{Lusch2018} implements concepts from Koopman Operator theory \cite{kutz2016koopman} for dynamical models, where the linearity of the Koopman Operator is exploited to advance in time the dynamical fields in a reduced space. The Sparse Identification of Nonlinear Dynamics (SINDy) is proposed in \cite{Brunton2016}, where the reduced vectors are assumed to follow a dynamics governed by a library of functions determined a priori.

Recently, \textit{Neural Operators} (NO) \cite{kovachki2021neural, bartolucci2023representation}, i.e., DL models whose objective is the \textit{approximation of operators} instead of functions - contrary to what is typical in DL - have found applications in surrogate modeling tasks. As in the case of PDEs we deal with a mapping between infinite-dimensional functional spaces (from the space of initial and boundary conditions to the solution space of the PDE), the approximated operator is called the \textit{solution operator} of the PDE. The (chronologically) first works on Neural Operators are the DeepONets \cite{Lu2021} and the Fourier Neural Operator \cite{li2020fourier}. Since these publications, the literature on NO has flourished, with many theoretical and empirical studies \cite{kovachki2021neural, CNO, Lu2022_afaircomparison, hao2023gnot, kissas2022learning,li2024geometry, gupta2022towards, jin2022mionet}. In some related works Graph Deep Learning has been used for surrogate modeling to generalize to different geometries \cite{brandstetter2022message,pichi2023graph, Franco2023, equer2023multi}. Beside DR, our model also leverages Neural ODEs \cite{chen2018neural} (NODEs), which are a class of Neural Networks (NNs) where the state of the system $h(t)\in\mathbb{R}^D$ behaves according to $\frac{dh(t)}{dt}=f_\theta(h(t),t)$, with $f_{\theta}$ being parametrized by a NN.
NODEs present the advantage of modeling the dynamics of $h(t)$ continuously in time.
\subsection{Related works}
Among the large literature on Neural Operators and methods at the intersection between DL and ROM, the papers closest to our work are: \cite{knigge2024space} where the latent dynamics is described by an ODE, \cite{yin2023continuous} where the latent dynamics is required to respect the equivariant properties of the original PDE, \cite{vcnef-hagnberger:2024} where an architecture featuring Neural Fields and Transformers is used to solve PDEs for different initial conditions and PDE parameters and \cite{takamoto2023learning} where a general method for including the PDE parameter dependency into Neural Operators is proposed.

\subsection{Contributions}
In this work we propose an autoregressive DL-based method for solving \textbf{parametrized}, \textbf{time-dependent} and (typically) \textbf{nonlinear} PDEs exploiting \textit{dimensionality reduction} and \textit{Neural ODEs}. Our contributions are the following:
\begin{itemize}
    \item We construct a model that allows for the variation of both the PDE's \textbf{parameters} and \textbf{initial conditions}. 
    \item We define two mappings parametrized by $2$ NNs: a close-to bijective mapping between the \textbf{full} (high-fidelity) PDE solution space and the \textbf{latent} (low-fidelity) space via an AutoEncoder (AE) and a mapping from the latent vector at time $t_i$ to the next latent vector at time $t_{i+1}$ modeled by a (latent) NODE. 
    \item Training on a \textbf{given} $\Delta t$ we show that our model can generalize at testing time to \textbf{finer} time steps $\Delta t'<\Delta t$. 
    \item We show a simple but effective strategy to train this model combining a \textbf{Teacher Forcing} type of training with an approach which takes into account the \textbf{Autoregressive} nature of this model at testing time. 
    \item We achieve computational speed up compared to standard numerical PDE solvers thanks to $3$ factors: doing inference at a $\Delta t$ \textbf{higher} than what is usually required by standard numerical solvers, solving an \textbf{ODE} instead of a PDE and advancing in time in a \textbf{low} dimensional space instead of the full original space.
    \item We test our methodology on a series of PDEs benchmarks (using \cite{takamoto2022pdebench} among others) and show that thanks to DR, our model is (at least 2 times) \textbf{lighter} and (at least 2 times) \textbf{faster} than current State of the Art (SOTA) methods.
\end{itemize}
 
\section{Methodology}
\label{sec:methodology}
\subsection{Mathematical background}
Let $\mathcal{V}$, $\mathcal{S}$ and $\mathcal{S}^0$ be the functional spaces of the boundary condition functions, PDE solution functions and the initial condition functions of a given PDE, respectively. These functions are required to satisfy some properties, such that:\begin{equation}
 \left\{
\begin{aligned}
& \mathcal{V} = \{v|v:\partial\mathcal{D}_{\mathbf{x}}\times\mathcal{D}_t\rightarrow\mathbb{R}^m\,;\lVert v_i \rVert_2<\infty\,\forall\,i\in[1,\cdots,m]\,\,; v\in\mathcal{C}^2\},\\
&  \mathcal{S} = \{s|s:\mathcal{D}_{\mathbf{x}}\times\mathcal{D}_t\rightarrow\mathbb{R}^m\,;\lVert s_i \rVert_2<\infty\,\forall\, i \,\in[1,\cdots,m]\,\,; s\in\mathcal{C}^2\},\\
& \mathcal{S}^0\subseteq \mathcal{S}^t = \{s(x,t=\tilde{t}|\pmb{\mu})|s(x,t=\tilde{t}|\pmb{\mu}):\mathcal{D}_{\mathbf{x}}\rightarrow\mathbb{R}^m, \forall \tilde{t}\in \mathcal{D}_t, \pmb{\mu}\in \mathcal{D}_{\pmb{\mu}}\},
\end{aligned}
\right.
\end{equation}
where $\mathcal{S}^t$ is the set of all possible states, $ \mathcal{D}_{\mathbf{x}}\subseteq\mathbb{R}^n$ is the domain of independent variables $\mathbf{x}$, $\mathcal{D}_{t}\subseteq\mathbb{R}^+$ is the temporal domain and $\partial\mathcal{D}_{\mathbf{x}}\subseteq\mathbb{R}^n$ the boundary of $\mathcal{D}_{\mathbf{x}}$ and $\mathcal{D}_{\pmb{\mu}}\subseteq\mathbb{R}^z$ is a domain for the vector $\pmb{\mu} = (\mu_1,\mu_2,\cdots, \mu_z)$, containing information about the PDE parameters, the geometry of the problem and whatever quantity defines the physical system. 
We are interested in solving general PDEs of the form of:
\begin{equation}
    \left\{
    \begin{aligned}
    &\hat{\mathcal{N}}(s(\mathbf{x},t|\pmb{\mu}),\mathbf{x},t,\pmb{\mu})= g(\mathbf{x},t,\pmb{\mu} ) \\
    &s(\tilde{\mathbf{x}},t|\pmb{\mu}) = v(\tilde{\mathbf{x}},t,\pmb{\mu}) \\
    &s(\mathbf{x},t=0|\pmb{\mu}) = s^0(\mathbf{x},\pmb{\mu}),
    \end{aligned}
    \right.
\label{eq:PDEsystem}
\end{equation}
where $\hat{\mathcal{N}}(s(\mathbf{x},t|\pmb{\mu}),\mathbf{x},t,\pmb{\mu})$ is a (typically) nonlinear integro-differential operator, $g(\mathbf{x}, t ,\pmb{\mu})$ is the forcing term, $s\in\mathcal{S}$ is the PDE solution, $v\in\mathcal{V}$ and $s^0\in\mathcal{S}^0$ are the boundary and initial conditions, $\mathbf{x}\in\mathcal{D}_{\mathbf{x}}$, $ \pmb{\mu}\in\mathcal{D}_{\pmb{\mu}}$, $\tilde{\mathbf{x}}\in\partial\mathcal{D}_{\mathbf{x}}$, $ t\in\mathcal{D}_{t}$. The different elements of the system of equations (\ref{eq:PDEsystem}) have either an explicit dependency on $\pmb{\mu}$, signaled by $(\cdot,\pmb{\mu})$, or an implicit dependency, signaled by $(\cdot|\pmb{\mu})$. Although (\ref{eq:PDEsystem}) describes a very general PDE system and our method description addresses this fully general case, in the experiments shown in Section \ref{sec:results} we fix $v(\tilde{\mathbf{x}},t, \pmb{\mu})$, i.e., it is not an input to the NN, we choose $g(\mathbf{x},t,\pmb{\mu} ) = 0$ and $\hat{\mathcal{N}} = \hat{\mathcal{N}}(s(\mathbf{x},t|\pmb{\mu}),\mathbf{x},\pmb{\mu})$, i.e., without explicit $t$ dependence.

In the context of surrogate modeling for parametric PDEs, one usually approximates by means of a NN either the \textit{Solution Operator} $\hat{\mathcal{S}}$ (\textbf{global} approach) or the \textit{Evolution Operator} $\hat{\mathcal{H}}$ (\textbf{autoregressive} approach), where 
\begin{equation}
\hat{\mathcal{S}}:\mathcal{S}^0\times\mathcal{D}_{t}\times\mathcal{D}_{\pmb{\mu}}\rightarrow\mathcal{S}\quad\text{and}\quad \hat{\mathcal{H}}:\mathcal{S}^t\times\mathcal{D}_{\Delta t}\times\mathcal{D}_{\pmb{\mu}}\rightarrow\mathcal{S}^t,
\label{eq:global&auto_op}
\end{equation}
with $\mathcal{D}_{\Delta t}\subseteq\mathbb{R}^+$. When approximating $\hat{\mathcal{S}}$, the NN is given as input $(s^0,t,\pmb{\mu})$ to output $s(\mathbf{x},t|\pmb{\mu})$, while when approximating $\hat{\mathcal{H}}$, the NN is given as input $(s(\mathbf{x},t=\tilde{t}|\pmb{\mu}), \Delta t, \pmb{\mu})$ to output $s(\mathbf{x},\tilde{t}+\Delta t|\pmb{\mu})$. While the former can approximate the solution $s$ at any point in time $t$ with just one call of the solver, the latter requires advancing iteratively in time by predicting the solution at the next time step from the solution at the previous time step as input, starting from $s^0$. Although the global approach has a (potential) advantage in terms of computational speed, we propose an autoregressive method for the following reasons:
\begin{itemize}
    \item Most PDEs represent causal physical phenomena, hence their solution evolution at time $t$ only depends on the system state at $t$. Therefore, as it is done in classical numerical solvers, only the solution $s$ at time $t$ is necessary for the prediction of $s$ at time $t+\Delta t$. This fact is not respected by global approaches.
    \item Global approaches require in general a high number of NN weights, as a mapping for arbitrary $t$ is required, contrary to autoregressive methods, as the state $s(t)$ carries more information than $s^0$ to predict $s(t+\Delta t)$.
\end{itemize}
While the two approaches are clearly different from a theoretical perspective, from a purely \textit{architectural} point of view they are very similar, as an architecture conceived as global can always be used autoregressively and vice-versa. It is primarily the training strategy that determines whether a global or autoregressive logic is followed.

In what follows we show how to approximate with a NN the (latent) Evolution Operator that governs the dynamics of the \textbf{reduced} space to which the full space $\mathcal{S}$ is mapped.
\subsection{Discretization}
In order to work with solution functions $s$ computationally it is necessary
to discretize the independent variable (typically spatial at least), temporal and parametric domains.  We thus define:
$\mathbf{X} = \{x_k|x_k\in \mathcal{D}_{\mathbf{x}}, x_k = (x_k^1,\cdots,x_k^n),k=0,\cdots,N_\mathbf{x}\}$ as the set of points inside the domain of independent variables; 
$\partial\mathbf{X} = \{\tilde{x}_k|\tilde{x}_k\in \partial\mathcal{D}_{\mathbf{x}}, \tilde{x}_k = (\tilde{x}_k^1,\cdots,\tilde{x}_k^n), k=0,\cdots,N_{\tilde{\mathbf{x}}}\}$ as the set of points on the boundary of the \textit{independent} variables of the domain $\mathcal{D}_{\mathbf{x}}$;
$\pmb{\mathcal{X}} = \mathbf{X}\cup\partial\mathbf{X}$,
$\pmb{M} = \{\pmb{\mu}\in\mathcal{D}_{\pmb{\mu}},\pmb{\mu}=(\mu_0,\mu_1,\cdots,\mu_z)\}$ as the set of parameter points $\pmb{\mu}$;
$\mathbf{T} = \{t|t\in\mathcal{D}_{t},t=(t_0,t_1,\cdots,t_{F})\}$  as the set of discrete points in time.
We also define the \textit{solution} and the \textit{initial condition} sets, as the sets of functions living in $\mathcal{S}$ and $\mathcal{S}^0$ discretized on $\pmb{\mathcal{X}}$ and $\pmb{T}$:
$\mathcal{S}_r=\{s_{r}(\mathbf{x},t|\pmb{\mu})|\mathbf{x}\in\pmb{\mathcal{X}}, t\in\mathbf{T}, \pmb{\mu}\in\pmb{M}\}\subset{\mathbb{R}^{|\pmb{\mathcal{X}}|\times n\times m}}$ and $\mathcal{S}^{0}_r=\{s^0_{r}(\mathbf{x},\pmb{\mu})|\mathbf{x}\in\pmb{\mathcal{X}},\pmb{\mu}\in\pmb{M}\}\subset\mathcal{S}_r$, where $r$ is a subscript that indicates a discretized representation of $s$. Obviously, $s_r(\mathbf{x},t_0|\pmb{\mu}) = s_r^0(\mathbf{x},\pmb{\mu})$. We will signal the implicit parameter dependency of $s_r$ using the notation $s_r(\mathbf{x},t|\pmb{\mu})$. In principle $s_r(\mathbf{x},t|\pmb{\mu},s_r^0)$ but for notational ease we will drop the implicit dependence on $s_r^0$. Although we are using a finite difference approach for discretization, our methodology is fully general to other discretization schemes too (finite volumes, finite elements, etc.)
\subsection{Reduced space and (latent) Neural ODEs}
\label{subsec:AE&NODE}
We want to build a method that at inference time maps the initial condition $s^0$ into its reduced representation and then evolves it in time (according to the PDE parameters) autoregressively.
The \textbf{first} building block of our methodology is the mapping between the full and the reduced space by an AutoEncoder. Let $\mathcal{E}$ be the \textit{reduced} (\textit{latent}) \textit{set}
\begin{equation}
    \mathcal{E}=\{\varepsilon(t|\pmb{\mu})|\varepsilon(t|\pmb{\mu})=
    (\varepsilon_1(t|\pmb{\mu}), \cdots,\varepsilon_{\lambda}(t|\pmb{\mu})), t\in\mathcal{D}_t, \pmb{\mu}\in\mathcal{D}_{\pmb{\mu}}\}\subset{\mathbb{R}^{\lambda}},
\end{equation}
with $\lambda\ll|\mathcal{X}|\cdot nm$ being the dimension of the latent space. Each time-dependent vector $\varepsilon(t|\pmb{\mu})\in\mathcal{E}$ has a one-to-one correspondence with a given solution function $s\in\mathcal{S}$ (implicitly depending on the parameter $\pmb{\mu}$), so that by computing the dynamics of $\varepsilon(t|\pmb{\mu})$ we can reconstruct the original trajectory of $s(\mathbf{x},t|\pmb{\mu})$. Each dimension $\varepsilon_i(t)$ is an \textit{intrinsic representation} of the corresponding function $s$ and embodies the correlations, symmetries and fundamental information about the original object $s$ (for a deeper understanding of the nature and the desiderata of a latent representation, see \cite{Eastwood2018AFF, Higgins2018TowardsAD}). Although we will work with discretized functions belonging to $\mathcal{S}_r$, each vector $\varepsilon(t|\pmb{\mu})$ is in principle associated with the original \textbf{continuous} function belonging to $\mathcal{S}$ (i.e., $\varepsilon(t|\pmb{\mu})$ should be independent of the discretization of $\mathcal{S}$).

The mathematical operators mapping $\mathcal{S}$ to $\mathcal{E}$ and viceversa are the \textit{Encoder} $\varphi$ and the \textit{Decoder} $\psi$, such that:
\begin{equation}
    \varphi: \mathcal{S}\rightarrow\mathcal{E}\quad\text{and}\quad\psi: \mathcal{E}\rightarrow\mathcal{S},
    \label{eq:AE}
\end{equation}
with $\varphi\circ\psi=\psi\circ\varphi= \mathbbm{1}$, together forming the AutoEncoder. We approximate $\varphi$ and $\psi$ by two NNs, respectively $\varphi_\theta:\mathcal{S}_r\rightarrow\mathcal{E}$ and $\psi_\theta:\mathcal{E}\rightarrow\mathcal{S}_r$.
The \textbf{second} building block concerns the dynamics of the vectors $\varepsilon$ belonging to the reduced set $\mathcal{E}$. We assume that the temporal dynamics of $\mathcal{E}$ follows an ODE:
\begin{equation}
    \frac{d}{dt}\varepsilon(t|\pmb{\mu})=f(\varepsilon(t|\pmb{\mu}),\pmb{\mu}),\quad\text\quad f\in\mathcal{F}:\mathcal{E}\times\mathcal{D}_{\pmb{\mu}}\rightarrow\mathcal{E},
    \label{eq:ODE}
\end{equation}
where $\pmb{\mu}\in\pmb{M}$ is the vector of PDE parameters.$f$ does not depend \textit{explicitly} on $t$ since the PDEs we work with do not have explicit time dependence, making the dynamics of $\mathcal{E}$ an \textit{autonomous system}. If instead $\hat{\mathcal{N}}$ or $g(\mathbf{x},t,\pmb{\mu})$ had an explicit dependence on $t$, we would have $f = f(\varepsilon(t|\pmb{\mu}),\pmb{\mu},t)$ and would treat $t$ in the model simply as an additional dimension of $\pmb{\mu}$. We can now define the \textit{Processor}
\begin{equation}
\label{eq:processor}
    \pi:\mathcal{E}\times\mathcal{F}\times\mathcal{D}_{\pmb{\mu}}\times\mathcal{D}_{\Delta t}\rightarrow \mathcal{E},
\end{equation} as the mathematical operator that advances the latent vector $\varepsilon(t|\pmb{\mu})$ in time according to Equation (\ref{eq:ODE}):
\begin{equation}
    \pi(\varepsilon(t_i|\pmb{\mu}),f,\pmb{\mu},\Delta t_{i+1,i}) = \varepsilon(t_i|\pmb{\mu})+\int_{t_i}^{t_{i+1}}f(\varepsilon(t|\pmb{\mu}),\pmb{\mu}) \,dt,
    \label{eq:NODE}
\end{equation}
with $\Delta t_{i+1,i} = t_{i+1}-t_i$ and the $\pmb{\mu}$ dependency being controlled by $f$. Clearly, $\pi(\varepsilon(t_i|\pmb{\mu}),f,\pmb{\mu},\Delta t_{i+1,i})= \varepsilon(t_{i+1}|\pmb{\mu})$. In summary, $\varphi$ and $\psi$ describe the mapping between the full order and reduced order representation of the system, while $\pi$ describes the \textit{dynamics} of the system. For notational convenience, we will \textbf{drop} the dependence of $\pi$ on $f$.

We now define $f_\theta$ as a NN which approximates $f$ and $\pi_\theta$ as the discrete approximation of $\pi$ which advances in time the vectors belonging to $\mathcal{E}$ by solving the integral of Equation (\ref{eq:NODE}), using known integration schemes (see in~\ref{sec:RK}):
\begin{equation}
   \pi_\theta(\varepsilon(t_{i}|\pmb{\mu}), \pmb{\mu},\Delta t_{i+1,i}) = ODESolve(\varepsilon(t_{i}|\pmb{\mu}),\pmb{\mu},\Delta t_{i+1,i}),
\end{equation}
as it is done in Neural ODEs (NODEs) \cite{chen2018neural,Kidger2022OnND}.
For example, in the case of the explicit Euler scheme \cite{Ascher1998ComputerMF}:
\begin{equation}
   \pi_\theta(\varepsilon(t_{i}|\pmb{\mu}),\pmb{\mu},\Delta t_{i+1,i}) = \varepsilon(t_i|\pmb{\mu})+\Delta t_{i+1,i}\,f_\theta(\varepsilon(t_{i}|\pmb{\mu}),\pmb{\mu}).
\end{equation}
The Processor $\pi$ is the equivalent of the Evolution Operator $\hat{\mathcal{H}}$ of Equation (\ref{eq:global&auto_op}) but acting on the reduced space $\mathcal{E}$ of \textbf{discrete} intrinsic representations: as such $\pi$ does not need to be equipped with the notion of (spatial) discretization invariance as in the case of Neural Operators. 
\begin{figure*}
  \centering
  \includegraphics[width=1.0\textwidth]{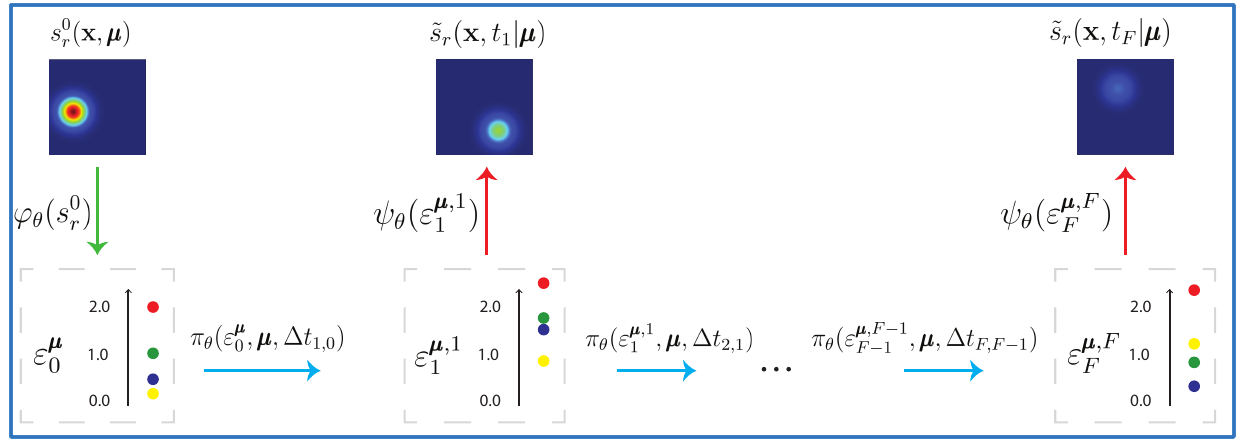}
  \caption{Workings of our proposed method at \textbf{testing} time. The initial condition $s^0_r$ is mapped trough the Encoder $\varphi_\theta$ into its latent representation $\varepsilon_0^{\pmb{\mu}}$. Subsequently the vector $\varepsilon_0^{\pmb{\mu}}$ is advanced in time autoregressively by repeated evaluation of the processor $\pi_\theta$, conditioned to the vector of parameters $\pmb{\mu}$ and to the size of the temporal jump $\Delta t_{i+i,i}$. The Decoder $\psi_\theta$ is used to map back each predicted latent vector $\varepsilon_{i}^{\pmb{\mu},i}$ into the corresponding field $\tilde{s}_r(\mathbf{x},t_i|\pmb{\mu})$. Notice that $\varphi_\theta$ is applied only to the initial condition $s_r^0$.}
  \label{fig:inference}
\end{figure*}
\subsection{Training of the model}
\label{subsec:training}
The model we defined in Section \ref{subsec:AE&NODE} requires the optimization of \textbf{two} training processes which we consider \textbf{coupled}: the training of the AE which regulates the mappings between $\mathcal{S}_r$ and $\mathcal{E}$ and the training of $\pi_\theta$ which regulates the latent dynamics described by Equation (\ref{eq:ODE}). The latter can be approached by combining a \textit{Teacher Forcing (TF)} and an \textit{Autoregressive (AR)} strategy. We thus define: 
\begin{equation}
    \mathcal{L}_{1,i} = \frac{||s_r(\mathbf{x},t_i|\pmb{\mu})-\psi_\theta\circ\varphi_\theta(s_r(\mathbf{x},t_i|\pmb{\mu})||_2}{||s_r(\mathbf{x},t_i|\pmb{\mu})||_2}
\end{equation}
as the term which governs the AE training. By introducing
\begin{equation}
    \left\{
    \begin{aligned}
    &\varepsilon_i^{\pmb{\mu}} = \varphi_\theta(s(\mathbf{x},t_i|\pmb{\mu})),\\
    &\varepsilon_{i}^{\pmb{\mu},k} = \pi_\theta(\cdot,\pmb{\mu},\Delta t_{i,i-1})\circ\cdots\circ\pi_\theta(\varepsilon_{i-k}^{\pmb{\mu}},\pmb{\mu},\Delta t_{i-k+1,i-k}), \\
    \end{aligned}
    \right.
\label{eq:notation_simplified}
\end{equation}
we define the two terms which govern the latent dynamics:
\begin{equation}
    \left\{
    \begin{aligned}
    &\mathcal{L}_{2,i}^{T,k_1} = \frac{||\varepsilon_i^{\pmb{\mu}}-\varepsilon_{i}^{\pmb{\mu},k_1}||_2}{||\varepsilon_i^{\pmb{\mu}}||_2},\\
    &\mathcal{L}_{2,i}^{A,k_2} =  \frac{||\varepsilon_i^{\pmb{\mu}}-\varepsilon_i^{\pmb{\mu},i}||_2}{||\varepsilon_i^{\pmb{\mu}}||_2},
    \end{aligned}
    \right.
\label{eq:L2_loss}
\end{equation}
where $T$ identifies the TF approach and $A$ the AR one. The term $\mathcal{L}_{2,i}^{T,k_1}$ (TF), penalizes the difference between the expected latent vector $\varepsilon_{i}^{\pmb{\mu}}$ and the predicted latent vector $\varepsilon_{i}^{\pmb{\mu},k_1}$ obtained by applying autoregressively $\pi_\theta$ to $\varepsilon_{i-k_1}^{\pmb{\mu}}$, \textit{which comes from encoding} the \textbf{true} \textit{field} $s_r(\mathbf{x},t_{i-k_1}|\pmb{\mu})$ (hence the name \textit{Teacher-Forcing}, as the \textbf{true} input $k_1$ steps earlier, is fed into the NN). Using TF when training autoregressive models is known to cause potential \textit{distribution shift} \cite{brandstetter2022message}, representing a problem at inference time: as depicted in Figure \ref{fig:inference}, at testing time the input of $\pi_\theta$ is the previous output of $\pi_\theta$ starting from $\varepsilon_0^{\pmb{\mu}}$, contrary to what $\mathcal{L}^{T,k_1}_{2,i}$ penalizes (unless $k_1= F$, i.e., the full length of the time series). To avoid this mismatch between training and inference, we introduced $\mathcal{L}^{A,k_2}_{2,i}$ (AR), which penalizes the difference between the expected latent vector $\varepsilon_{i}^{\pmb{\mu}}$ and the predicted latent vector $\varepsilon_i^{\pmb{\mu},i} = \pi_\theta(\cdot,\pmb{\mu},\Delta t_{i,i-1})\circ\cdots\circ\pi_\theta(\varepsilon_{0}^{\pmb{\mu}},\pmb{\mu},\Delta t_{1,0})$ obtained by repeated application of $\pi_\theta$ \textit{starting from the encoded representation of the initial condition} $s_r^0$, as it is done at testing time. $k_2$ denotes the number of steps in time from which the \textbf{gradients of the backpropagation algorithm} flow, i.e., the predicted latent vector at time $t_i$ is obtained by encoding the initial condition $s_r^0$ and fully evolving it autoregressively (by applying $\pi_\theta$ $i$ times), but the gradients of the backpropagation algorithm \textbf{flow only} from the predicted latent vector at time $t_{i-k_2}$ up to $t_i$. It follows that $\mathcal{L}^{T,k_1}_{2,i}$ and $\mathcal{L}^{A,k_2}_{2,i}$ are computed in the same way only if $k_1=k_2= F$. By truncating the gradients flow at time $t_{i-k}$, we are implementing a form of Truncated Backpropagation Through Time (TBPTT) as it is usually done for gradients stability purpuses when training Recurrent Neural Networks (RNNs) \cite{pmlr-v115-aicher20a}.
Figure \ref{fig:inference} shows a summary of the method at testing time, where the predicted solution is computed as
$\tilde{s}_r(\mathbf{x},t_i|\pmb{\mu}) =
     \psi_\theta\circ \pi_\theta(\cdot,\pmb{\mu},\Delta t_{i,i-1})\circ\cdots\circ\pi_\theta(\cdot,\pmb{\mu},\Delta t_{1,0})\circ\varphi_\theta(s^0_r)$.
\subsection{Combining Teacher Forcing with Autoregressive}
\label{subsec:TF_&_AR}
\textbf{How} do we combine AR and TF strategies in practice? We start by considering the loss $\mathcal{L}^{T,1}_{2,i}$, the simplest form of TF strategy with the advantage of being computationally efficient and stable, but at the cost of making the training agnostic to the autoregressive nature of the model at testing time and not addressing the \textit{accumulation of errors} which is typical of autoregressive models. $\mathcal{L}^{A,k_2}_{2,i}$ instead, regardless of the chosen $k_2$, already reflects during training the autoregressive modality used at testing; it has however the disadvantage of being computationally more demanding (for $k_2>1$) and more difficult to train the larger $k_2$ is. If $k_1>1$, $\mathcal{L}^{T,k_1}_{2,i}$ introduces a certain degree of 'autoregressiveness' as well, although the latent vector at time $t_{i-k}$ is still provided by the true solution. Among the several possible training strategies, here we list the two we used, with $\mathcal{L}_{2,i}=\beta \mathcal{L}^{T,k_1}_{2,i}+\gamma \mathcal{L}^{A,k_2}_{2,i}$, where $\beta$ and $\gamma$ weigh the importance of the terms:
\begin{enumerate}
    \item set $\beta=1$, $\gamma=0$ and $k_1=1$, using only the TF term.
    \item set $\beta=1$, $\gamma=1$, $k_1=1$ and dynamically increase $k_2$ during the training, starting with $k_2 = 1$. This strategy has the advantage of taking into account the AR term gradually during the training.
\end{enumerate}
Our experiments have shown that although for some systems strategy $1$ is enough, more complex datasets require using strategy $2$, mainly due to two separate behaviors in our observations. First, that in the \textbf{early stages} of the training, $\mathcal{L}_{1,i}$, $\mathcal{L}^{T,1}_{2,i}$ and $\mathcal{L}^{A,1}_{2,i}$ play the important role of building a latent space whose dynamics is described by Equation (\ref{eq:ODE}); and second, that in the \textbf{later stages} of the training, with $k_2$ becoming larger (and the computed gradients more complex), the autoregressive nature of the model is increasingly taken into account, with $\pi_\theta$ becoming more robust to the accumulation of errors.
\subsection{Generalization in the time domain}
\label{subsec:generalization_in_time}

\begin{figure*}
  \centering
  \includegraphics[width=0.8\textwidth]{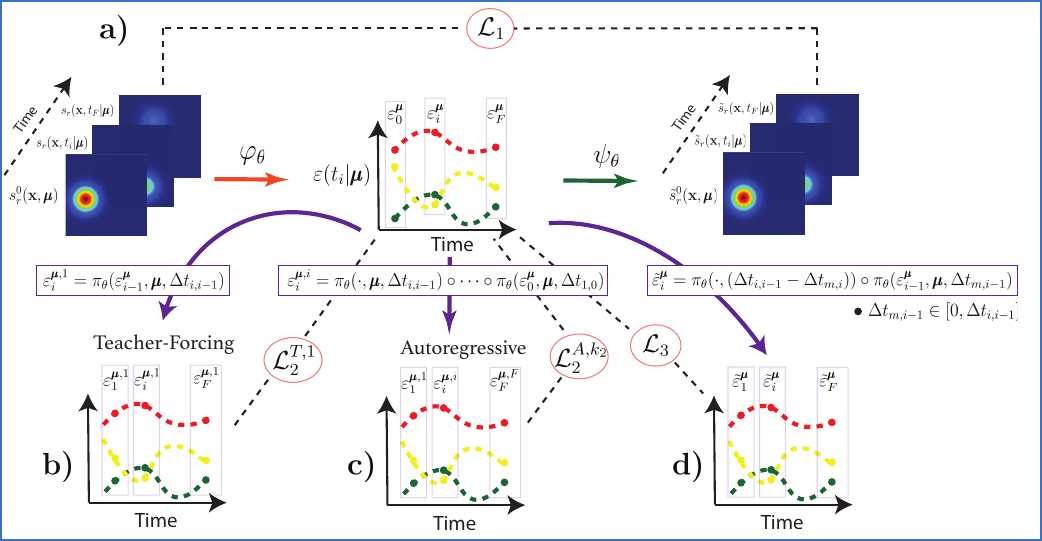}
  \caption{A representation of the \textbf{training} procedure. \textbf{a)} The time series of fields $s_r(\mathbf{x},t_i|\pmb{\mu})$, with $i\in\{0,F\}$, is processed by the Encoder $\varphi_\theta$ and the corresponding latent vectors $\varepsilon_i^{\pmb{\mu}}$ are obtained; these are subsequently mapped back to the full space by means of the Decoder $\psi_\theta$ which generates the time series of reconstructed fields $\tilde{s}_r(\mathbf{x},t_i|\pmb{\mu})$, allowing for the computation of $\mathcal{L}_1$. \textbf{b)} The Processor $\pi_\theta$ receives as input the sequence of latent vectors $\varepsilon_i^{\pmb{\mu}}$ with $i\in\{0,F-1\}$ and predicts the latent vectors $\varepsilon_{i}^{\pmb{\mu},1}$ with $i\in\{1,F\}$. $\mathcal{L}_2^{T,1}$, where $T$ stands for \textit{Teacher-Forcing}, is thus computed with inputs $\varepsilon_i^{\pmb{\mu}}$ and $\varepsilon_{i}^{\pmb{\mu},1}$. \textbf{c)} The Processor $\pi_\theta$ is applied autoregressively to the initial latent vector $\varepsilon_0^{\pmb{\mu}}$ and the whole time series of vectors $\varepsilon_{i}^{\pmb{\mu},i}$ is reconstructed with $i\in\{1,F\}$; $\mathcal{L}_2^{A,k_2}$, where $A$ stands for \textit{Autoregressive}, is thus computed with inputs $\varepsilon_i^{\pmb{\mu}}$ and $\varepsilon_{i}^{\pmb{\mu},i}$. \textbf{d)} The Processor $\pi_\theta$ takes as input the sequence of latent vectors $\varepsilon_i^{\pmb{\mu}}$ with $i\in\{0,F-1\}$ and outputs for each $\varepsilon_i^{\pmb{\mu}}$ an intermediate vector $\varepsilon^{\pmb{\mu},1}_{m}$ with a time-step $\Delta t_{m,i-1}$ randomly sampled from $[0,\Delta t_{i,i-1}]$. Last, $\pi_\theta$ advances in time each $\varepsilon^{\pmb{\mu},1}_{m}$ with a time-step of $\Delta t_{i,i-1}-\Delta t_{m,i}$ to get the predicted vectors $\tilde{\varepsilon}_{i}^{\pmb{\mu}}$; $\mathcal{L}_3$ is thus computed with inputs $\varepsilon_i^{\pmb{\mu}}$ and $\tilde{\varepsilon}_{i}^{\pmb{\mu}}$.}
  \label{fig:architecture}
\end{figure*}

As shown in Figure \ref{fig:inference_in_time} we expect our models to be trained on a given set of time-steps, but we want them to generalize to time-steps not seen during the training phase (such as intermediate times). For this reason, we introduce a \textbf{last term} of the loss function as:
\begin{equation}
\label{eq:L3}
    \left\{
    \begin{aligned}
    &\mathcal{L}_{3,i} =\frac{||\varepsilon_i^{\pmb{\mu}}-\tilde{\varepsilon}_{i}^{\pmb{\mu}}||_2}{||\varepsilon_i^{\pmb{\mu}}||_2}\ \\ &\tilde{\varepsilon}_{i}^{\pmb{\mu}} = \pi_\theta(\cdot,(\Delta t_{i,i-1}-\Delta t_{m,i}))\circ\pi_\theta(\varepsilon_{i-1}^{\pmb{\mu}},\pmb{\mu},\Delta t_{m,i-1}),
    \end{aligned}
    \right.
\end{equation}
where $\Delta t_{m,i-1}\in[0,\Delta t_{i,i-1}]$ is a randomly sampled intermediate time step and $i-1<m<i$. In Appendix \ref{subsubsec:RK_time_gen} we further detail $\mathcal{L}_{3,i}$. In some cases we also found it beneficial to add a \textit{regularization} term $\mathcal{L}_{rg}$ to the latent vectors, such as $\mathcal{L}_{rg} = \lambda_{rg}\sum_{i=0}^F ||\varepsilon_i^{\pmb{\mu}}||_1/\lambda$, with $\lambda_{rg}\in\mathbb{R}^+$. 

Thus, during model training for a given $s_r^0(\mathbf{x},\pmb{\mu})$, the gradients are computed based on the final loss function of:
\begin{equation}
    \begin{aligned}
    &\mathcal{L}_{tr} = \frac{1}{F}\sum_{i=0}^F \alpha \mathcal{L}_{1,i}+\frac{1}{F}\sum_{i=1}^{F}\left[\beta\mathcal{L}_{2,i}^{T,k_1}+\gamma\mathcal{L}_{2,i}^{A,k_2}+\delta\mathcal{L}_{3,i}\right]+ \mathcal{L}_{rg} = \alpha \mathcal{L}_{1}+\beta\mathcal{L}_{2}^{T,k_1}+\gamma\mathcal{L}_{2}^{A,k_2}+\delta\mathcal{L}_{3} + \mathcal{L}_{rg},
    \end{aligned}
\end{equation}
where $k_1$ and $k_2$ depend on the chosen strategy of section \ref{subsec:TF_&_AR} and $\alpha$, $\beta$, $\gamma$ and $\delta$ weigh the importance of each term.  We use $\mathcal{L}_{tr}$ for training and $\mathcal{L}_{vl}$ for validation:
\begin{equation}
    \mathcal{L}_{vl} = \mathcal{L}_{tr} + \sum_{i=1}^{F} \frac{||s_r(\mathbf{x},t_i|\pmb{\mu}) - \tilde{s}_r(\mathbf{x},t_i|\pmb{\mu})||_2}{||s_r(\mathbf{x},t_i|\pmb{\mu})||_2}.
\end{equation}
Figure \ref{fig:architecture} visualizes our training methodology.

\section{Results}
\label{sec:results}
In this section we compare our method with a series of SOTA methods from \cite{vcnef-hagnberger:2024} and \cite{takamoto2022pdebench}. The datasets we use for comparison are taken from \cite{takamoto2022pdebench}. A complete description of the PDEs can be found in Appendix \ref{subsec:datasets}. In Appendix \ref{subsec:training_details} we list all the training and hyperparameter details and in Appendix \ref{sec:methods_comparison} the methods used for comparison. We use as metric the nRMSE error defined in Equation (\ref{eq:nRMSE}). 
\definecolor{lightgray}{gray}{0.9} 

\begin{table}[]
\begin{center}
\begin{tabular}{ll>{\columncolor{lightgray}}ll}
\toprule
PDE   & Model & nRMSE, $\Delta t = 0.05\,s$    & nRMSE, $\Delta t = 0.01\,s$ \\
\hline
& (Ours) & $\mathbf{0.0066}$&$\mathbf{0.0066}$ \\
& FNO & $0.0190$ & $0.0258$\\
& MP-PDE & $0.0195$& \\
& UNet & $0.0079$ & \\
1D Advection& CORAL & $0.0198$&$0.9656$  \\
& Galerkin & $0.0621$ &\\
& OFormer & $0.0118$& \\
& VCNeF & $0.0165$&$0.0165$ \\
& VCNeF-R & $0.0113$& \\ \hline
& (Ours) & $\mathbf{0.0373}$&$\mathbf{0.0399}$ \\
& FNO & $0.0987$ &$0.1154$\\
& MP-PDE & $0.3046$ &\\
& UNet & $0.0566$& \\
1D Burgers & CORAL &$0.2221$&$0.6186$  \\
& Galerkin & $0.1651$& \\
& OFormer & $0.1035$& \\
& VCNeF & $0.0824$&$0.0831$ \\
& VCNeF-R & $0.0784$& \\ 
 \bottomrule
\end{tabular}
\end{center}
\caption{nRMSE on test dataset for fixed $\pmb{\mu}$ and varying $s^0$ for the 1D Advection and Burgers datasets. Gray column refers to testing with the $\Delta t$ of the training, while white one with a smaller $\Delta t$. Cells are empty when comparison was not found in literature.}
\label{table:results_fixed_par}
\end{table}
\begin{table}[]
\begin{center}
\begin{tabular}{ll>{\columncolor{lightgray}}l>{\columncolor{lightgray}}ll}
\toprule
PDE   & Model & nRMSE, $\Delta t = 0.05\,s$    & nRMSE,$\Delta t = 0.01\,s$ & nRMSE,$\Delta t = 0.01\,s$ \\
\hline
& (Ours) & $0.0028$&&$\mathbf{0.0032}$ \\
& FNO & &$0.0044$ &\\
2D SW & U-Net &  &$0.0830$ &\\
 & PINN&&$0.0170$ &\\ 
 \bottomrule
\end{tabular}
\end{center}
\caption{nRMSE on test dataset for fixed $\pmb{\mu}$ and varying $s^0$ for the 2D Shallow-Water dataset. Gray column refers to testing with the $\Delta t$ of the training, while white one with a smaller $\Delta t$. Our model is trained on $\Delta t = 0.05\,s$, while the others on $\Delta t = 0.01\,s$. }
\label{table:results_fixed_par_SW}
\end{table}
\begin{figure*}
  \centering
  \includegraphics[width=0.9\textwidth]{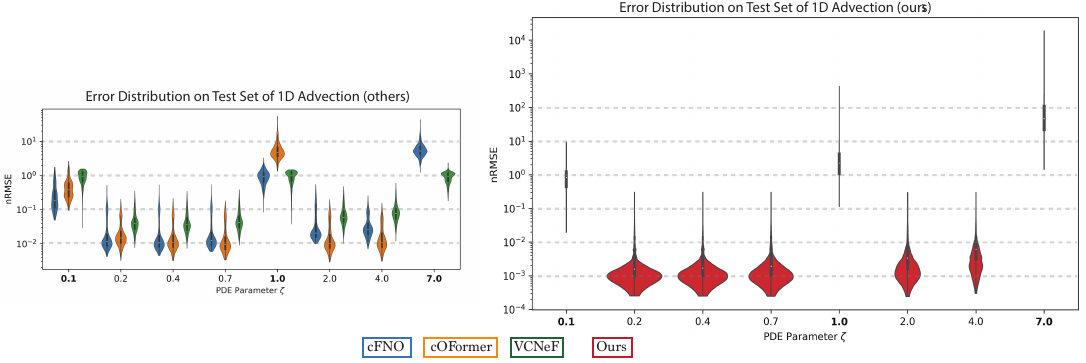}
  \caption{Distribution of the nRMSE across the test sample for the parametric 1D Advection. Regular font on the x axes refers to training parameter values, while bald ones to testing parameters (but in both cases testing initial conditions). We compare our methodology (red on right image) with other published methods (left image, taken from \cite{vcnef-hagnberger:2024}).}
  \label{fig:advection_parametric}
\end{figure*}
\subsection{PDEs with fixed parameter}
\label{subsec:PDE_fixed}
In this Section we are going to apply our method to the 1D Advection Equation (\ref{eq:advection}) ($\zeta =0.1$), to the 1D Burgers' Equation (\ref{eq:burger}) ($\nu=0.001$) and to the 2D Shallow-Water (SW) Equations (\ref{eq:sw}).
In Table \ref{table:results_fixed_par} and \ref{table:results_fixed_par_SW} we compare our results to the ones obtained (on the same dataset), in \cite{takamoto2022pdebench} and \cite{vcnef-hagnberger:2024}. In Table \ref{table:results_fixed_par} we show that our proposed model achieves a \textbf{lower nRMSE} compared to a series of common methods used in Scientific Machine Learning for the $3$ cases. Furthermore, we observe that our model achieves a better \textbf{generalization in time} than the other methods in the Burgers' and Advection cases, meaning that we got little to none increase of the nRMSE when going from testing on the training $\Delta t=0.05\,s$ to testing on a smaller $\Delta t = 0.01\,s$ . For the SW case in Table \ref{table:results_fixed_par_SW}, even if our model is trained with $\Delta t = 0.05\,s$ while the others with $\Delta t = 0.01\,s$, we still get a lower nRMSE when testing on $\Delta t = 0.01\,s$ (thus the comparison on the same number of time-steps is only between our method in the white column and the other methods in the grey column). For the experiments in this section we used Strategy $1$ of Section \ref{subsec:TF_&_AR}, as using $\mathcal{L}_{2,i}^{T,1}$ alone was sufficient to reach acceptable results.
\begin{figure*}
  \centering
  \includegraphics[width=0.9\textwidth]{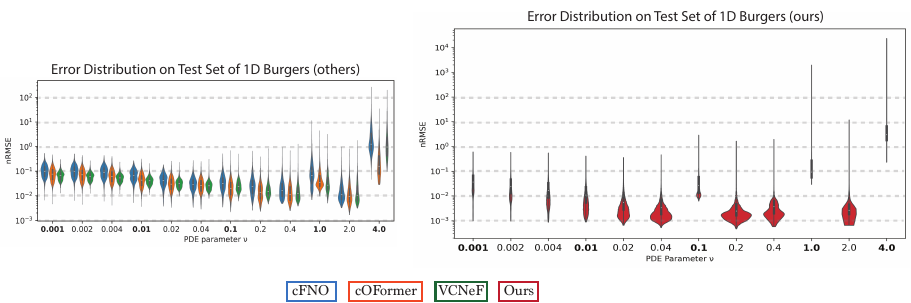}
  \caption{Distribution of the nRMSE across the test sample for the parametric 1D Burgers'. Regular font on x axes refers to training parameters, while bald ones to testing parameters (but in both cases testing initial conditions). We compare our methodology (red on right image) with other published methods (left image, taken from \cite{vcnef-hagnberger:2024}).}
  \label{fig:burger_parametric}
\end{figure*}
\subsection{PDEs with varying parameters}
\label{subsec:PDE_varying}
\begin{figure*}
  \centering
  \includegraphics[width=0.8\textwidth]{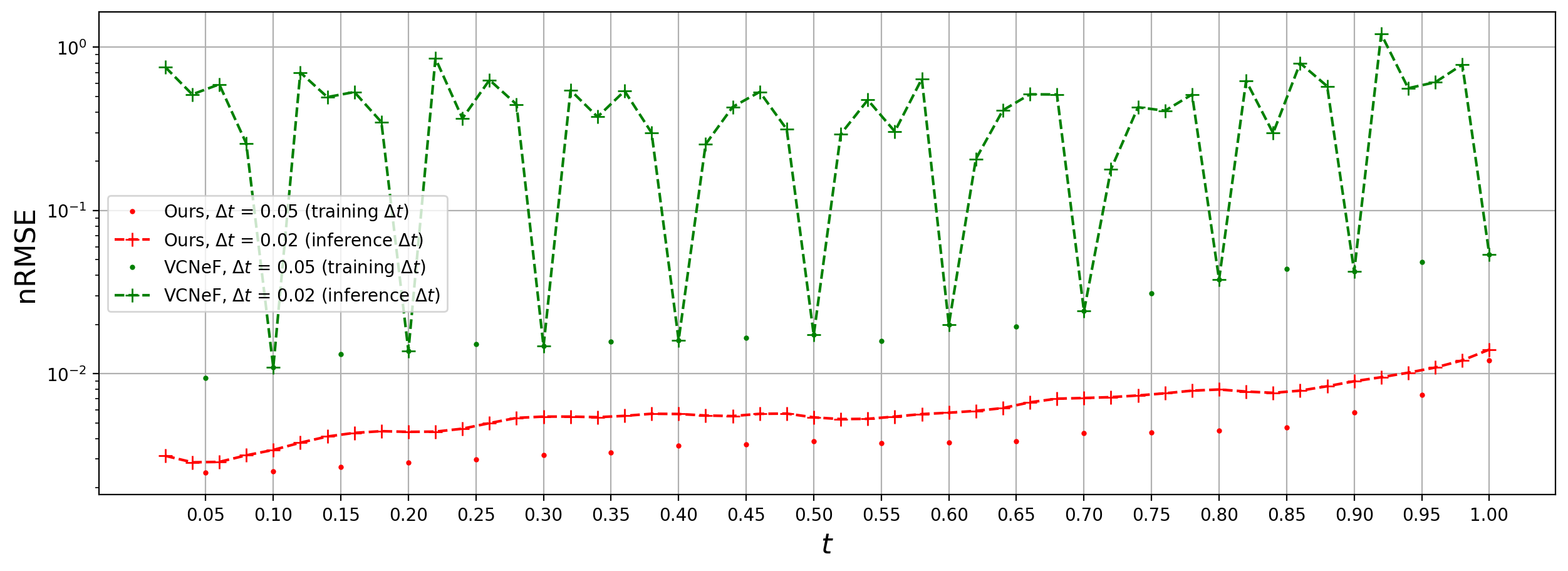}
  \caption{Comparison (on the test dataset) for the Molenkamp application of the nRMSE over time $t$ between our model (red) and the VCNeF (green). We study the difference when applying at inference the same $\Delta t $ used for the training ($\Delta t = 0.05\,s$) and when applying a smaller one $\Delta t = 0.02\,s$. The nRMSE of our model slightly increases when decreasing the $\Delta t $, while VCNeF struggles with inference at intermediate time-steps.}
  \label{fig:molenkamp}
\end{figure*}

In this section we experiment with $3$ datasets where we both vary the initial conditions and the PDE parameters: 1D Advection Equation (\ref{eq:advection}), the 1D Burgers' Equation (\ref{eq:burger}) and the 2D Molenkamp Test (\ref{eq:molenkamp}). In all three cases we use Strategy $2$ of Section \ref{subsec:TF_&_AR}, since only using $\mathcal{L}_{2,i}^{T,1}$ optimized correctly $\mathcal{L}_{tr}$ but resulted in a larger value of $\mathcal{L}_{vl}$. We start with $k_2 =1$ and we increase it by $1$ every $30$ epochs until the maximum length of the time series is reached. To make the training more stable, we introduce gradually the $\gamma\,\mathcal{L}_{2,i}^{A,k_2}$ term by starting with a $\gamma = \gamma_0<1$ and increasing it every epoch by an amount of $\gamma_0$ until $\gamma=1$. In Figures \ref{fig:advection_parametric} and \ref{fig:burger_parametric} we show a comparison of our methodology (red) with the cFNO, cOFormer and VCNeF from \cite{vcnef-hagnberger:2024}. In both cases, we show the distribution across the test samples of training (regular font) and testing (bald font) parameters when using \textbf{testing initial conditions}. From Figure \ref{fig:advection_parametric}, we see that our model has a lower median than the compared methods on the training velocities $\zeta$, while it struggles with testing parameters, similar to cFNO, cOFormer and VCNeF too. This is likely a dataset issue yielding insufficient generalization, with $0.1$ and $7.0$ both being outside the training range and $1.0$ possessing a dynamics very far from the one that characterizes $0.7$ and $2.0$. Similarly for the Burgers' case in Figure \ref{fig:burger_parametric}, the median of the nRMSE given by our model is lower than the compared methods for all parameters $\nu$ except $\nu=1.0$ and $\nu=4.0$. In this case our model - similar to the ones used for comparison - is able to generalize better than in the Advection example to test parameters, as in the case of $\nu=0.001$ and $\nu=0.01$. Given the discrepancy in the ability of the models to generalize to different testing parameters, more accurate strategies for adaptively selecting parameter points for training should be researched. In Figure \ref{fig:molenkamp} we compare our method with VCNeF on the Molenkamp when testing on $\Delta t = 0.05\,s$ (same as the one in training) and when $\Delta t = 0.02\,s$: our method achieves a lower nRMSE and is able to generalize to intermediate time points better than VCNeF. In Appendix \ref{subsec:speed_and_numb_comp}, we compare the number of parameters and the inference speed of the methods used and we show that our proposed method is \textbf{lighter} and \textbf{faster} at inference.
\subsection{Ablation studies}
In Appendix \ref{subsec:solver_for_time} we conduct ablation studies regarding how the choice of the ODE solver and $\mathcal{L}_{3,i}$ impacts the capabilities of the method and the generalization in time.

\section{Conclusions}
In this work we showed how \textbf{Dimensionality Reduction} and \textbf{Neural ODEs} can be coupled to construct a surrogate model of time-dependent and parametric PDEs. Our model inherits from these two paradigms two important features which are desiderata when building DL models that substitute standard numerical solvers, i.e.,  \textbf{fast computational inference} and \textbf{continuity in time}. The former is achieved thanks to the low dimensionality of the reduced space $\mathcal{E}$, while the latter by the definition of the latent dynamics through the ODE of Equation \ref{eq:ODE}. In Section \ref{sec:results} we showed that our methodology \textbf{surpasses in accuracy} several state of the art methods on different benchmarks used in the Scientific Machine Learning field. In addition, our model requires \textbf{significantly less NN's weights} (thus less memory) and is \textbf{computationally faster at inference} compared to other published methodologies (Tables \ref{tab:inference_time} and \ref{tab:numb_weights}); for these reasons relying on \textit{dimensionality reduction} as opposed to \textit{large and overparametrized} architectures is going to be key in the future for building fast and memory efficient surrogate models of complex physical systems. 


The main limitation of our method is the use of CNNs for Encoding and Decoding, which hinders its applicability to non-uniform meshes and makes it necessary to re-train models if inference needs to be done on grid points not used at training - future research will explore using Neural Operators for the construction of the Encoder and the Decoder. Another aspect which should be improved concerns the definition of an efficient strategy to determine which PDE parameters should be used during training in order to be able to better generalize to new ones, as evident from Section \ref{subsec:PDE_varying}. Finally, while leaving the construction of $\mathcal{E}$ and definition of the function $f_\theta$ general gives flexibility to the fitting of the training dataset, researching into the \textit{interpretability} of both $\mathcal{E}$ and $f_\theta$ can at the same time improve our understanding of NNs and build more accurate surrogate models.

\subsection{CRediT authorship contribution statement}
\textbf{Alessandro Longhi}: Conceptualization, Methodology, Software, Validation, Formal analysis, Investigation, Data Curation, Writing - Original Draft, Visualization. \\
\textbf{Danny Lathouwers}: Conceptualization, Methodology, Formal Analysis, Resources, Supervision, Project administration, Funding acquisition.\\
\textbf{Zoltán Perkó}: Conceptualization, Methodology, Formal analysis, Resources, Writing - Review and Editing, Supervision, Project administration, Funding acquisition.

\newpage
\appendix
\section*{Appendix}
\section{Runge-Kutta schemes}
\label{sec:RK}
Runge-Kutta methods \cite{Ascher1998ComputerMF} are a family of numerical methods for the solution of Ordinary Differential Equations (ODEs). They belong to the category of \textit{one-step} methods, as such they do not use any information from previous time steps. Given $t_i\in\mathbf{T}$, a standard explicit Runge-Kutta method would solve Equation (\ref{eq:ODE}) as:
\begin{equation}
\label{eq:rungekutta_reduced}
    \varepsilon(t_{i+1}|\pmb{\mu}) = \varepsilon(t_{i}|\pmb{\mu}) + \Delta t_{i+1,i}\sum_{j=1}^q h_j b_j,
\end{equation}
where $q$ is called the \textit{stage} of the Runge-Kutta approximation, $\Delta t_{i+1,i} = t_{i+1}-t_i$ and:
\begin{align*}
     b_1 = &f(\varepsilon(t_{i}|\pmb{\mu}),t_{i},\pmb{\mu}),\\
     b_2 = &f(\varepsilon(t_{i}|\pmb{\mu})+(a_{2,1}b_1)\Delta t_{i+1,i},t_i+c_2\Delta t_{i+1,i},\pmb{\mu}),\\
    & \vdots \\
    b_k = & f(\varepsilon(t_{i}|\pmb{\mu})+\sum_{l=1}^{k-1}a_{k,l}b_l\Delta t_{i+1,i}, t_i+c_k\Delta t_{i+1,i},\pmb{\mu}).
\end{align*}
The matrix composed by $a_{ij}$ is known as Runge-Kutta matrix, $h_j$ are the weights and $c_j$ are the nodes, with their values given by the Butcher tableau \cite{butcher_1963}.

\subsection{The effect of the stage of RK on time generalization}
\label{subsubsec:RK_time_gen}
\begin{figure}
  \centering
  \includegraphics[width=0.8\textwidth]{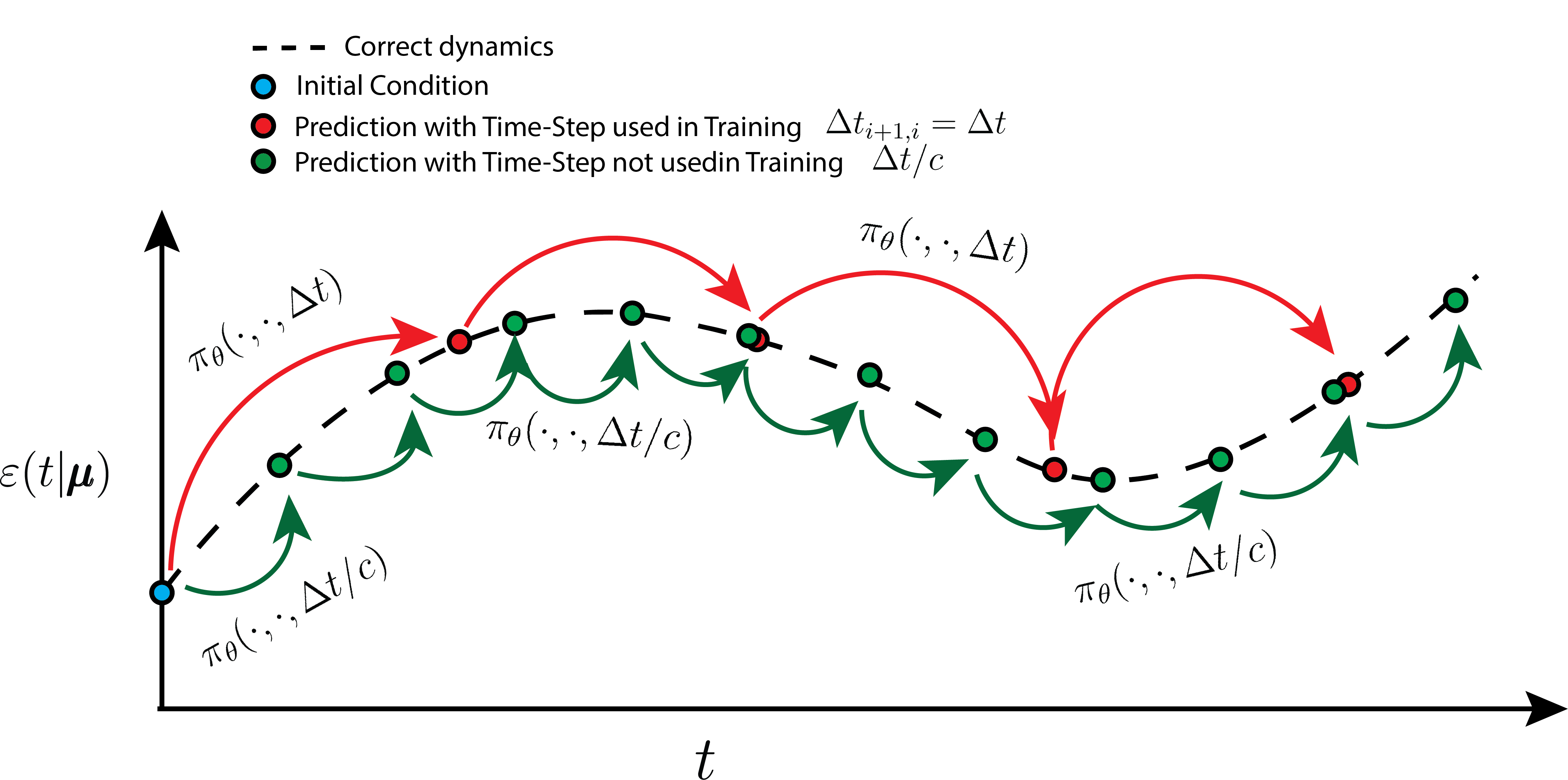}
  \caption{The evolution over time of a one dimensional $\varepsilon(t|\pmb{\mu})$ is shown (dot line). The red points indicate the steps in time used during the training, at intervals of $\Delta t_{i+1,1}$, and the green points show the points in time that can be predicted at testing time at distance of $\Delta t_{i+1,1}/\alpha$, where $\alpha\in[1,\infty)$.}
  \label{fig:inference_in_time}
\end{figure}

In this work we use a fixed Runge-Kutta time stepper, and to go from the state $i$ to the state $i+1$ we do not step trough intermediate states. We defined the Processor of Equation (\ref{eq:processor}) with a $\Delta t_{i+1,i}$ dependency as we want to perform inference even at temporal discretizations finer than the one used at training. 
Although this task may look trivial as $\pi_\theta$ directly takes $\Delta t_{i+1,i} $ as input, it raises the following issue when solving the ODE. Let us consider a processor $\pi_\theta$ which evolves in time the latent vector $\varepsilon(t|\pmb{\mu})$ using an Euler integration scheme (from here on we omit the $\pmb{\mu}$ dependency for ease of reading):
\begin{equation}
    \varepsilon(t_{i+1})  = \varepsilon(t_{i})+\Delta t_{i+1,i}\,f_\theta(\varepsilon(t_{i})),
\end{equation}
and let us define a moment in time $t_m$ such that $t_i<t_m<t_{i+1}$. We want $f_\theta$ to satisfy the following conditions:
\begin{equation}
    \left\{
    \begin{aligned}
    &\varepsilon(t_{i+1})  = \varepsilon(t_{i})+\Delta t_{i+1,i}\,f_\theta(\varepsilon(t_{i})), \\
    &\varepsilon(t_{i+1})  = \varepsilon(t_{m}) + (\Delta t_{i+1,i}-\Delta t_{m,i})\, f_\theta(\varepsilon(t_{m}))),
    \end{aligned}
    \right.
\label{eq:two_generalization}
\end{equation}
where $\varepsilon(t_{m}) = \varepsilon(t_{i})+\Delta t_{m,i}\,f_\theta(\varepsilon(t_{i}))$.
By taking the difference of the two Equations of (\ref{eq:two_generalization}) we get that 

\begin{equation}
f_\theta(\varepsilon(t_{i})) = f_\theta(\varepsilon(t_{m})),
\end{equation}
i.e., when we use the Euler scheme $f_\theta$ must be a constant if we want $\pi_\theta$ to be coherent with its predictions at the variation of $\Delta t_{i+1,i}$. Notice that a constant $f_\theta$ would imply a linear time dependence of $\varepsilon(t)$ (the dotline of Figure \ref{fig:inference_in_time} would be a line), meaning that the construction of $\mathcal{E}$ would be subjected to a \textbf{strong constraint}, thus limiting the expressiveness of the AE. For a RK method of order 2 instead:
\begin{equation}
    \varepsilon(t_{i+1}) = \varepsilon(t_{i})+\Delta t_{i+1,i}f_\theta\left(\varepsilon(t_{i})+\frac{1}{2}\Delta t_{i+1,i}\,f_\theta(\varepsilon(t_{i}))\right),
\end{equation}
where $f_\theta$ is evaluated at $\Delta t_{i+1,i}/2$. We want $f_\theta$ to respect the following system:
\begin{equation}
    \left\{
    \begin{aligned}
    &\varepsilon(t_{i+1}|\pmb{\mu})  = \varepsilon(t_{i}|\pmb{\mu}) + \Delta t_{i+1,i}f_\theta(\varepsilon(t^{i+1}_i|\pmb{\mu})), \\
    &\varepsilon(t_{i+1}|\pmb{\mu})= \varepsilon(t_{i}|\pmb{\mu})+\Delta t_{m,i} f_\theta(\varepsilon(t^m_i|\pmb{\mu}))+ (\Delta t_{i+1,i}-\Delta t_m)f_\theta(\varepsilon(t^{i+1}_m|\pmb{\mu})),
    \end{aligned}
    \right.
\label{eq:two_generalization_RK_2}
\end{equation}
where  $t_i<t_m<t_{i+1}$, $t^{j}_k = \frac{t_j+t_k}{2}$ and $\varepsilon(t^{j}_k|\pmb{\mu}) =\varepsilon(t_{k}|\pmb{\mu})+\frac{ \Delta t_{j,k}}{2}f_\theta(\varepsilon(t_{k}|\pmb{\mu}))$. By taking the difference of System (\ref{eq:two_generalization_RK_2}) we get:
\begin{equation}
    \Delta t_{i+1,i}f_\theta(\varepsilon(t^{i+1}_i|\pmb{\mu})) = \Delta t_{m,i} f_\theta(\varepsilon(t^m_i|\pmb{\mu}))+ (\Delta t_{i+1,i}-\Delta t_m)f_\theta(\varepsilon(t^{i+1}_m|\pmb{\mu})),
\end{equation}
which, by going back to full notation, results in the following Equation:
\begin{equation}
    \begin{aligned}
    &\Delta t_{i+1,i}f_\theta\left[\varepsilon(t_{i}|\pmb{\mu})+\frac{ \Delta t_{i+1,i}}{2}f_\theta(\varepsilon(t_{i}|\pmb{\mu}))\right]= \Delta t_{m,i} f_\theta\left[\varepsilon(t_{i}|\pmb{\mu})+\frac{ \Delta t_{m,i}}{2}f_\theta(\varepsilon(t_{i}|\pmb{\mu}))\right]+ \\ 
    &+(\Delta t_{i+1,i}-\Delta t_m)f_\theta\left[\varepsilon(t_{i}|\pmb{\mu})+\frac{ \Delta t_{m,i}}{2}f_\theta(\varepsilon(t_{i}|\pmb{\mu}))+\frac{ \Delta t_{i+1,m}}{2}f_\theta\left[\varepsilon(t_{i}|\pmb{\mu})+\frac{ \Delta t_{m,i}}{2}f_\theta(\varepsilon(t_{i}|\pmb{\mu}))\right]\right],
    \end{aligned}
\end{equation}
where $\varepsilon(t_{i}|\pmb{\mu})+\frac{ \Delta t_{m,i}}{2}f_\theta(\varepsilon(t_{i}|\pmb{\mu})) = \varepsilon(t_m|\pmb{\mu})$. The constraint to which $f_\theta$ is now subjected allows for a more complex form of $f_\theta$ which in turns results in a reduced space $\mathcal{E}$ \textbf{more capable of adapting} to the complexity of the original space $\mathcal{S}$. It follows that, if we require $f_\theta$ to generalize to a variable $\Delta t_{i+1,i}$, the higher the stage of the Runge-Kutta scheme used, the more complex $f_\theta$ can be and the more complex the reduced space $\mathcal{E}$ can be.

\section{Architecture details}
\label{subsec:architecture}
As detailed in Section \ref{subsec:AE&NODE}, our model is made up of three components: an Encoder $\varphi_\theta$, a Processor $\pi_\theta$ and a Decoder $\psi_\theta$.\\

\textbf{Encoder}\\
We build $\varphi_\theta$ as a series of Convolutional layers \cite{CNNarithme}  followed by a final Linear layer as in the 2D example of Table \ref{tab:Encoder}.
\begin{table}[]
\begin{tabular}{|l|l|l|l|l|l|}
\hline
Layer Number & Input size & Output size & Filters & Kernel & Stride \\ \hline
$1$ (Convolutional) & $[m,N,N] $   & $[Fe_1,N,N]$                    & $Fe_1$ & $[Ke_{1},Ke_{1}]$ & $[1,1]$ \\ \hline
$2$ (Convolutional) & $[Fe_1,N,N]$ & $[Fe_2,\frac{N}{2},\frac{N}{2}]$ &  $Fe_2$    &   $[Ke_{2},Ke_{2}]$    &   $[2,2]$    \\ \hline
$\vdots$ &            &                                &      &       &                   \\ \hline
 $j$ (Convolutional)&  $[Fe_{j-1},\frac{N}{2^{j-2}},\frac{N}{2^{j-2}}]$ & $[Fe_{j},\frac{N}{2^{j-1}},\frac{N}{2^{j-1}}]$ &  $Fe_j$&  $[Ke_{j},Ke_{j}]$    &  $[2,2]$      \\ \hline
 $\vdots$ &            &                                &      &       &               \\ \hline
$L-1$  (Flat layer) & & & & &    \\ \hline
$L$ (Linear) &$[Fe_{L-2}\times\frac{N}{2^{L-3}}\times\frac{N}{2^{L-3}}]$ &$[\lambda]$ & & &   \\ \hline
\end{tabular}
\caption{The structure of the Encoder $\varphi_\theta$ layer by layer, for a 2D case. $m$ is the dimension of the solution field $s(\mathbf{x},t|\pmb{\mu})$, $N$ is the size of the spatial discretization of the field in the $x$ and in the $y$ axis, $Fe = [Fe_1,\cdots,Fe_{L-2}]$ is the vector of convolutional filters, $Ke = [Ke_1,\cdots,Ke_{L-2}]$ is the vector of kernels , $j\in\{2,L-2\}$ and $\lambda$ is the latent dimension. The Flat layer takes all the features coming from the last Convolutional layer and flattens them in a 1D vector. This example is easily reduced to the 1D case.}
\label{tab:Encoder}
\end{table}
The first layer has stride $1$ to do a preprocessing of the fields and the subsequent layers up to the Flat layer halve each spatial dimension by $2$. We use as activation function after each Convolutional layer the GELU function \cite{hendrycks2016gaussian} and we \textbf{do not} use any activation function after the final Linear layer to not constrain the values of the latent space. We experimented with Batch Normalization \cite{ioffe2015batch} and Layer Normalization layers \cite{ba2016layer} between the Convolutional layers and the GELU function but we did not notice any improvements in the results. The weights of all the layers are initialized with the Kaiming (uniform) initialization \cite{he2015delving}. Notice that we are not using any Pooling layer \cite{gholamalinezhad2020pooling} to reduce the dimensionality but only strided Convolutions, as Pooling layers would enforce translational invariance which is not always a desired property.

\textbf{Processor}\\
Inside the Processor $\pi_\theta$, in practice only the function $f_\theta$ which approximates $f$ of Equation (\ref{eq:ODE}) is parametrized by a NN with $f_\theta(\varepsilon(t|\pmb{\mu}),\pmb{\mu})$ being a function of both $\varepsilon(t|\pmb{\mu})$ and $\pmb{\mu}$. We experimented with the parameter dependency in two ways:
\begin{itemize}
    \item $\pmb{\mu}$ is simply concatenated to the reduced vector $\varepsilon(t|\pmb{\mu})$. In this case $f_\theta:\mathbb{R}^{\lambda+z}\rightarrow\mathbb{R}^{\lambda}$, where $z$ is the dimensionality of $\pmb{\mu}$ and $\lambda$ the latent dimension. 
    \item The vector $\varepsilon(t|\pmb{\mu})$ is conditioned to $\pmb{\mu}$ through a FiLM layer \cite{perez2018film}. This means defining the function $\alpha:\mathbb{R}^z\rightarrow\mathbb{R}^{\lambda}$ and the function $\tau:\mathbb{R}^z\rightarrow\mathbb{R}^{\lambda}$. The input to $f_\theta$ will thus be $\alpha(\pmb{\mu})\odot\varepsilon(t|\pmb{\mu})$ + $\tau(\pmb{\mu})$, where $\odot$ is the Hadamard product. $\alpha$ and $\tau$ are chosen to be simple Linear layers.
\end{itemize}
In both cases $f_\theta$ is built as a sequence of Linear layers followed by the GELU activation function. Importantly the activation function is not used after the last Linear layer as this is a regression problem.

\textbf{Decoder}\\
\begin{table}[]
\begin{tabular}{|l|l|l|l|l|l|}
\hline
Layer Number & Input size & Output size & Filters & Kernel & Stride \\ \hline
$1$ (Linear) & $[\lambda] $   & $[Fd_{1}\times\frac{N}{2^{L-3}}\times\frac{N}{2^{L-3}}]$                    &  &  & \\ \hline
$2$ (Reshape layer) &  &  &     &      &      \\ \hline
$\vdots$ &            &                                &      &       &                   \\ \hline
 $j$ (T. Convolutional) & $[Fd_{j-2},\frac{N}{2^{L-j}},\frac{N}{2^{L-j}}]$ &  $[Fd_{j-1},\frac{N}{2^{L-j-1}},\frac{N}{2^{L-j-1}}]$ & $Fd_{j-1}$&  $[Kd_{j-1},Kd_{j-1}]$    &  $[2,2]$      \\ \hline
 $\vdots$ &            &                                &      &       &               \\ \hline
 $L-1$ (T. Convolutional) &$[Fd_{L-3},\frac{N}{2},\frac{N}{2}]$ &$[Fd_{L-2},N,N]$ &$Fd_{L-2}$ &$[Kd_{L-3},Kd_{L-3}]$ & $[2,2]$  \\ \hline
$L$ (T. Convolutional) &$[Fd_{L-2},N,N]$ &$[m,N,N]$ & $[m]$&$[Kd_{L-2},Kd_{L-2}]$ & $[1,1]$  \\ \hline
\end{tabular}
\caption{The structure of the Decoder $\psi_\theta$ layer by layer, for a 2D case. 'T.' stands for 'Transposed'.}
\label{tab:Decoder}
\end{table}
We build $\psi_\theta$ as a Linear layer followed by a series of Transposed Convolutional layers \cite{CNNarithme} as shown in Table \ref{tab:Decoder}. The initial Linear layer and the last Transposed Convolutional layer are \textbf{not} followed by an activation function while the other Transposed Convolutional layers are followed by a GELU function. We do not use any activation function for the Linear layer for symmetry with the Encoder, while for the last layer of the Decoder because this is a regression task. After the Reshape layer each Transposed (T.) Convolutional layer doubles the dimensionality in both $x$ and $y$ dimensions until the layer number $L-1$. The last layer does not increase the dimensionality of the input (stride $=1$) and is just used to go to the final dimensionality $m$ of the solution field $s$. $Fd = [Fd_1,\cdots,Fd_{L-2}]$ is the vector of convolutional filters, $Ke = [Kd_1,\cdots,Kd_{L-2}]$ is the vector of kernels, $j\in\{3,L-1\}$.
\subsection{Normalization of the inputs}
\label{subsec:normalization}
In order to facilitate the training process we normalize the inputs, as standard Deep Learning practice. We use a max-min normalization both for the input fields $s$ and for the parameters $\pmb{\mu}$. We do not normalize $\Delta t_{i+1,i}$. By max-min normalization, we mean the following: given an input $y$, we transform it accordingly to $y\rightarrow \frac{y-min(D_y)}{max(D_y)-min(D_y)}$, where $max(D_y)$ and $min(D_y)$ are computed over the \textit{training} datasets $D_y$ to which $y$ belongs. In our experiments $s$ is a scalar field so we only compute one tuple $(max(D_s),min(D_s))$. In the case of $\pmb{\mu}$ instead, since each parameter $\mu_i$ can belong to a different scale, we compute $z$ tuples $(max(D_{\mu_i}),min(D_{\mu_i}))$ with $i\in\{1,z\}$. We normalize accordingly the parameters for all the datasets while the input fields for all the datasets but the Burgers' Equation and the parametric Advection.

\section{Training and hyperparameter details}
\label{subsec:training_details}
In all the experiments we use the Adam optimizer \cite{adam} and we stop the training if the validation loss has not decreased for $200$ epochs. We use an Exponential Learning Rate Scheduler, with a decay parameter $\gamma_{lr}$. We set $5000$ as the maximum number of epochs. In all the experiments, unless otherwise specified, we used $q=4$ as the stage of the RK algorithm.  We use $\mathcal{L}_{tr}$ for training and $\mathcal{L}_{vl}$ for validation:
\begin{equation}
    \mathcal{L}_{vl} = \mathcal{L}_{tr} + \sum_{i=1}^{F} \frac{||s_r(\mathbf{x},t_i|\pmb{\mu}) - \tilde{s}_r(\mathbf{x},t_i|\pmb{\mu})||_2}{||s_r(\mathbf{x},t_i|\pmb{\mu})||_2}.
\end{equation}
\subsection{PDEs with fixed parameters}
\textbf{1D Advection}\\
We split the datasets  by taking the time series from the $1$st to the $8000$th as training, from the $8001$th to the $9000$th as validation and from the $9001$th to the $10000$th as testing. We use an initial learning rate of $0.001$, $\gamma_{lr} = 0.997$ and a batch size of $16$. We use $F_e =[8,16,32,64,64,64,64]$, $F_d =[64,64,64,64,32,16,1]$, $K_e = [5,5,5,5,5,5,5]$ and  $K_d = [6,6,6,6,6,6,5]$. $f_\theta$ is composed by $2$ hidden layers with $50$ neurons each and $\lambda=30$. $\lambda_{rg} = 0.0$\\
\textbf{1D Burgers}\\
We split the datasets  by taking the time series from the $1$st to the $8000$th as training, from the $8001$th to the $9000$th as validation and from the $9001$th to the $10000$th as testing. We use an initial learning rate of $0.0014$, $\gamma_{lr} = 0.999$ and a batch size of $32$. We use $F_e =[8, 16, 32, 32, 32, 32, 32]$, $F_d =[32, 32, 32, 32, 32, 16, 1,1]$, $K_e = [5,5,3,3,3,3,3]$ and  $K_d = [4,4,4,4,4,4,3,3]$. $f_\theta$ is composed by $4$ hidden layers with $200$ neurons each and $\lambda=30$. $\lambda_{rg} = 0.001$
\\
\textbf{2D Shallow-Water}\\
We split the datasets  by taking the time series from the $1$st to the $800$th as training, from the $801$th to the $900$th as validation and from the $901$th to the $1000$th as testing. We use an initial learning rate of $0.001$, $\gamma_{lr} = 0.999$ and a batch size of $16$. We use $F_e =[8, 32, 32, 32, 32, 32, 32]$, $F_d =[32, 32, 32, 32, 32, 16, 1,1]$, $K_e = [5,5,3,3,3,3,3]$ and  $K_d = [4,4,4,4,4,4,3,3]$. $f_\theta$ is composed by $2$ hidden layers with $50$ neurons each and $\lambda=20$. $\lambda_{rg} = 0.001$
\subsection{PDEs with varying parameters}
\textbf{1D Advection}\\
We use as training parameters the velocities $\zeta\in\{0.2,0.4,0.7,2.0,4.0\}$ and as testing $\zeta\in\{0.1,1.0,7.0\}$. We split the datasets  by taking, \textbf{for each training parameter} $\zeta$, the time series from the $1$st to the $8000$th as training, from the $8001$th to the $9000$th as validation and from the $9001$th to the $10000$th as testing. We use an initial learning rate of $0.0018$, $\gamma_{lr} = 0.995$ and a batch size of $64$. We use $F_e =[8,16,32,32,32,32,32]$, $F_d =[32,32,32,32,32,16,1]$, $K_e = [5,5,3,3,3,3,3]$ and  $K_d = [4,4,4,4,4,4,3]$. $f_\theta$ is composed by $4$ hidden layers with $200$ neurons each and $\lambda=30$. $\lambda_{rg} = 0.0$ and $\gamma_0=1/500$.\\
\textbf{1D Burgers}\\
We use as training parameters the diffusion coefficient $\nu\in\{0.002,0.004,0.02,0.04,0.2,0.4,2.0\}$ and as testing $\nu\in\{0.001,0.01,0.1,1.0,4.0\}$. We split the datasets  by taking the time series from the $1$st to the $8000$th as training, from the $8001$th to the $9000$th as validation and from the $9001$th to the $10000$th as testing. We use an initial learning rate of $0.0018$, $\gamma_{lr} = 0.995$ and a batch size of $124$. We use $F_e =[8,32,32,32,32,32,32]$, $F_d =[32,32,32,32,32,16,1,1]$, $K_e = [5,5,3,3,3,3,3]$ and  $K_d = [4,4,4,4,4,4,3,3]$. $f_\theta$ is composed by $4$ hidden layers with $200$ neurons each and $\lambda=30$. $\lambda_{rg} = 0.0$ and $\gamma_0=1/1000$.
\\
\textbf{2D Molenkamp test}\\
We generate $5000$ training, $200$ validation and $100$ testing samples by sampling from a uniform distribution the parameters described in $\ref{subsec:molenkamp_test}$. We use an initial learning rate of $0.0015$, $\gamma_{lr} = 0.995$ and a batch size of $16$. We use $F_e =[8, 16, 32, 32, 32, 32, 32]$, $F_d =[32, 32, 32, 32, 32, 16, 1, 1]$, $K_e = [5, 5, 3, 3, 3, 3, 3]$ and  $K_d = [4, 4, 4, 4, 4, 4, 3, 3]$. $f_\theta$ is composed by $2$ hidden layers with $100$ neurons each and $\lambda=50$. $\lambda_{rg} = 0.0$ and $\gamma_0=1/500$. Here we apply a GELU function  in the last layer of the Encoder and to the first layer of the Decoder. 
\subsection{Instabilities of the training}
\label{subsubsec:instabilities}
$\mathcal{L}_{2}^{T,k_1} $ and especially $\mathcal{L}_{2}^{A,k_2}$ can involve complex gradients. During the training, this can sometimes lead the NN to be stuck in the trivial minimum for $\mathcal{L}_{2,i}^{T} $ and $ \mathcal{L}_{2,i}^{A}$ which consists in $\varphi_\theta$ and $\pi_\theta$ returning a constant output. We found out that some datasets are particularly sensitive to this problem, while other are not affected by this issue. In the following, a list of measures which help avoiding the trivial solution:
\begin{itemize}
    \item removing the biases from the Encoder;
    \item using (Batch/Layer) Normalization layers in the Encoder (not necessarely after each convolution);
    \item careful tuning of the learning rate (lowering the learning rate or increasing the batch size);
    \item warm up of the learning rate;
    \item if the instabilities come mostly from $\mathcal{L}_{2,i}^{T,k_1} $ and $\mathcal{L}_{2,i}^{A,k_2} $ as they involve more complex gradients, it can be beneficial to turn them off for some initial epochs (for example during the warm up of the learning rate) by setting $\beta$ and $\gamma$ to zero. This allows for an initial construction of $\mathcal{E}$ with simpler constraints.
\end{itemize}
\subsection{Hardware details}
For training we use either an NVIDIA A40 40 GB or an NVIDIA A100 80GB PCIe depending on availabilities. 
\subsection{Number of NNs weights and speed of inference}
\label{subsec:speed_and_numb_comp}
In Table \ref{tab:numb_weights} we show the number of NNs weights associated with our model and the models used for comparison from \cite{vcnef-hagnberger:2024}. In Table \ref{tab:inference_time} we report the inference time for the Burgers' dataset with $\nu=0.001$. We do not consider the time spent for sending the batches from the CPU to the GPU; the time measured is the time taken to do inference on the whole testing dataset of 1000 initial conditions with a batchsize of 64. We use an NVIDIA A100 80GB PCIe to conduct the inference test. Inference time of other methods is from \cite{vcnef-hagnberger:2024} where they use an NVIDIA A100-SXM4 80GB GPU.

\begin{table}[]
\begin{tabular}{|l|l|l|}
\hline
\multicolumn{1}{|l|}{Time resolution} & Model    & Inference time {[}ms{]}  \\ \hline
                                      & Ours     & $466.73^{\pm 68.38}$                             \\ \cline{2-3} 
                                      & FNO      & $917.77^{\pm 2.51}$                 \\ \cline{2-3} 
\multicolumn{1}{|c|}{41}               & VCNeF    & $2244.04^{\pm 6.65}$                 \\ \cline{2-3} 
                                      & Galerkin & $2415.99^{\pm 54.56}$                  \\ \cline{2-3} 
                                      & VCNeF s. & $4853.17^{\pm 75.29}$                 \\ \cline{2-3} 
\multicolumn{1}{|l|}{}                & OFormer  & $6025.75^{\pm 12.75}$                 \\ \hline
                                      & Ours     & $932.43^{136.588}$                            \\ \cline{2-3} 
                                      & FNO      & $1912.19^{\pm 56.03}$                    \\ \cline{2-3} 
\multicolumn{1}{|c|}{81}               & VCNeF    & $4422.65^{\pm 4.11}$                  \\ \cline{2-3} 
                                      & Galerkin & $4940.80^{\pm 89.44}$                 \\ \cline{2-3} 
                                      & VCNeF s. & $9701.80^{\pm 84.48}$                  \\ \cline{2-3} 
\multicolumn{1}{|l|}{}                & OFormer  & $12081.98^{\pm 19.39}$                   \\ \hline
                                      & Ours     & $1440.67^{\pm 250.29}$                          \\ \cline{2-3} 
                                      & FNO      & $2808.04^{\pm 82.22}$                     \\ \cline{2-3} 
\multicolumn{1}{|c|}{121}               & VCNeF    & $6606.41^{\pm 3.0}$                  \\ \cline{2-3} 
                                      & Galerkin & $7908.18^{\pm 96.52}$                   \\ \cline{2-3} 
                                      & VCNeF s. & $14577.00^{\pm 112.83}$                    \\ \cline{2-3} 
\multicolumn{1}{|l|}{}                & OFormer  & $17965.47^{\pm 14.19}$                  \\ \hline
                                      & Ours     & $1846.729^{\pm 270.72}$                            \\ \cline{2-3} 
                                      & FNO      & $3733.10^{\pm 62.94}$                     \\ \cline{2-3} 
\multicolumn{1}{|c|}{161}               & VCNeF    & $6084.04^{\pm 9.37}$                    \\ \cline{2-3} 
                                      & Galerkin & $10295.78^{\pm 116.50}$                    \\ \cline{2-3} 
                                      & VCNeF s. & $19449.80^{\pm 113.73}$                  \\ \cline{2-3} 
\multicolumn{1}{|l|}{}                & OFormer  & $24108.24^{\pm 6.45}$                 \\ \hline
                                      & Ours     & $2389.07^{386.02}$                           \\ \cline{2-3} 
                                      & FNO      & $4614.21^{\pm 97.52}$                    \\ \cline{2-3} 
\multicolumn{1}{|c|}{201}               & VCNeF    & $7584.48^{\pm 1.86}$                  \\ \cline{2-3} 
                                      & Galerkin & $13151.47^{\pm 93.95}$                   \\ \cline{2-3} 
                                      & VCNeF s. & $24252.38^{\pm 101.41}$                   \\ \cline{2-3} 
\multicolumn{1}{|l|}{}                & OFormer  & $29986.81^{\pm 6.35}$                   \\ \hline
                                      & Ours     & $2773.019^{406.31}$                          \\ \cline{2-3} 
                                      & FNO      & $5572.07^{\pm 109.23}$                     \\ \cline{2-3} 
\multicolumn{1}{|c|}{240}               & VCNeF    & $8935.28^{\pm 7.08}$                  \\ \cline{2-3} 
                                      & Galerkin & $15600.60^{\pm 262.51}$                     \\ \cline{2-3} 
                                      & VCNeF s. & $29063.89^{\pm 79.58}$                 \\ \cline{2-3} 
\multicolumn{1}{|l|}{}                & OFormer  & $35900.51^{\pm 6.71}$                 \\ \hline
\end{tabular}
\caption{Comparison of the inference time. The dataset used is the Burger's one with $\nu=0.001$.}
\label{tab:inference_time}
\end{table}

\begin{table}[]
\begin{tabular}{llll}
\cline{2-4}
\multicolumn{1}{l|}{}                      & \multicolumn{3}{l|}{\# NNs weights }                                                                    \\ \hline
\multicolumn{1}{|l|}{Model}                & \multicolumn{1}{l|}{Advection}    & \multicolumn{1}{l|}{Burgers'} & \multicolumn{1}{l|}{Molenkamp} \\ \hline
\multicolumn{1}{|l|}{Ours}                 & \multicolumn{1}{l|}{188,549 / 214,461} & \multicolumn{1}{l|}{216,657}     & \multicolumn{1}{l|}{166,825}          \\ \hline
\multicolumn{1}{|l|}{Galerkin T.} & \multicolumn{1}{l|}{530,305}         & \multicolumn{1}{l|}{530,305}     & \multicolumn{1}{l|}{-}          \\ \hline
\multicolumn{1}{|l|}{FNO}                  & \multicolumn{1}{l|}{549,569}         & \multicolumn{1}{l|}{549,569}     & \multicolumn{1}{l|}{-}          \\ \hline
\multicolumn{1}{|l|}{U-Net}                & \multicolumn{1}{l|}{557,137}         & \multicolumn{1}{l|}{557,137}     & \multicolumn{1}{l|}{-}          \\ \hline
\multicolumn{1}{|l|}{MP-PDE}               & \multicolumn{1}{l|}{614,929}         & \multicolumn{1}{l|}{614,929}     & \multicolumn{1}{l|}{-}          \\ \hline
\multicolumn{1}{|l|}{OFormer}              & \multicolumn{1}{l|}{660,814}         & \multicolumn{1}{l|}{660,814}     & \multicolumn{1}{l|}{-}          \\ \hline
\multicolumn{1}{|l|}{VCNeF}                & \multicolumn{1}{l|}{793,825}         & \multicolumn{1}{l|}{793,825}     & \multicolumn{1}{l|}{1,594,005}          \\ \hline                         
\end{tabular}
\caption{Number of NNs' weights of the different architectures. In the first line of the 1D Advection case we show the number of weights of our model for $\zeta=0.01$/$\zeta$ varying.}
\label{tab:numb_weights}
\end{table}
\section{Methods used for comparison}
In Section \ref{sec:results} we compare our model to the following methods:
\\
\textbf{Fourier Neural Operator (FNO)} \cite{li2020fourier}: it is a particular case of a \textit{Neural Operator}, i.e., a class of models which approximate operators and that can thus perform mapping from infinite-dimensional spaces to infinite-dimensional spaces. The name comes from the the assumption that the Kernel of the operator layer is a convolution of two functions, which makes it possible to exploit the Fast Fourier Transform under particular circumstances.
\\
\textbf{Fourier Neural Operator (cFNO)} \cite{takamoto2023learning}: it is an adaptation of the FNO methodology which allows to add the PDE parameters as input.
\\
\textbf{Message Passing Neural PDE Solver (MP-PDE)} \cite{brandstetter2022message}: it leverages Graph Neural Networks (GNNs) for building surrogate models of PDEs. All the components are based on neural message passing which (as they show in the paper) representationally contain classical methods such as finite volumes, WENO schemes and finite differences.
\\
\textbf{U-Net} \cite{ronneberger2015u}: it is a method based on an Encoder-Decoder architecture with skip connections between the downsampling and upsampling procedures. Originally it was born for image segmentation and has been applied to the field of PDE solving \cite{gupta2022towards}.
\\
\textbf{Coordinate-based Model for Operator Learning (CORAL)} \cite{coral}: it is a method which leverages Neural Fields \cite{xie2022neural} for the solution of PDEs on general geometries and general time discretizations.
\\
\textbf{Galerkin Transformer (Galerkin)} \cite{cao2021choose}: it is a Neural Operator based on the self-attention mechanism from Transformers; it is based on a novel layer normalization scheme which mimics the Petrov-Galerkin projection. 
\\
\textbf{Operator Transformer (OFormer)} \cite{li2023transformer}: it is a Neural Operator which leverages the fact that the self-attention layer of Transformers is a special case of an Operator Layer as shown in \cite{kovachki2021neural} to build a PDE solver.
\\
\textbf{cOFormer}: it is an adaptation of the OFormer architecture which allows for the query of PDE parameters as inputs, following what is done in \cite{takamoto2022pdebench}. We 
\\
\textbf{Vectorized Conditional Neural Fields (VCNeF)} \cite{vcnef-hagnberger:2024}: it is a transformer based model which leverages neural fields to represent the solution of a PDE at any spatial point continuously in time. For the Molenkamp test we implemented the VCNeF method from the Git-Hub repository indicated in their paper: as they say in their paper, we use for the training of their model the One Cycle Scheduler with maximum learning rate at 0.2, initial division factor $0.003$ and final division factor $0.0001$; we use $500$ epochs, a batch size of $64$, an embedding size of $96$, $1$ transformer layer and $6$ modulation blocks. 
\label{sec:methods_comparison}
\\
\textbf{Physics-Informed Neural Networks (PINN)} \cite{pinns}: it is a class of methods which uses the physical knowledge of the system (in this case the PDE), in order to approximate the solution of the PDE. 
\\
The above methods have been implemented in \cite{vcnef-hagnberger:2024} (for results concerning 1D Advection and 1D Burgers) and \cite{takamoto2022pdebench} (for results concerning 2D Shallow Water) and we used their results as comparison with ours in Section \ref{sec:results}. 
\section{Ablation studies}
\subsection{The role of the ODE solver and of $\mathcal{L}_3$ in time generalization}
\label{subsec:solver_for_time}
\begin{figure}
  \centering
  \includegraphics[width=1.0\textwidth]{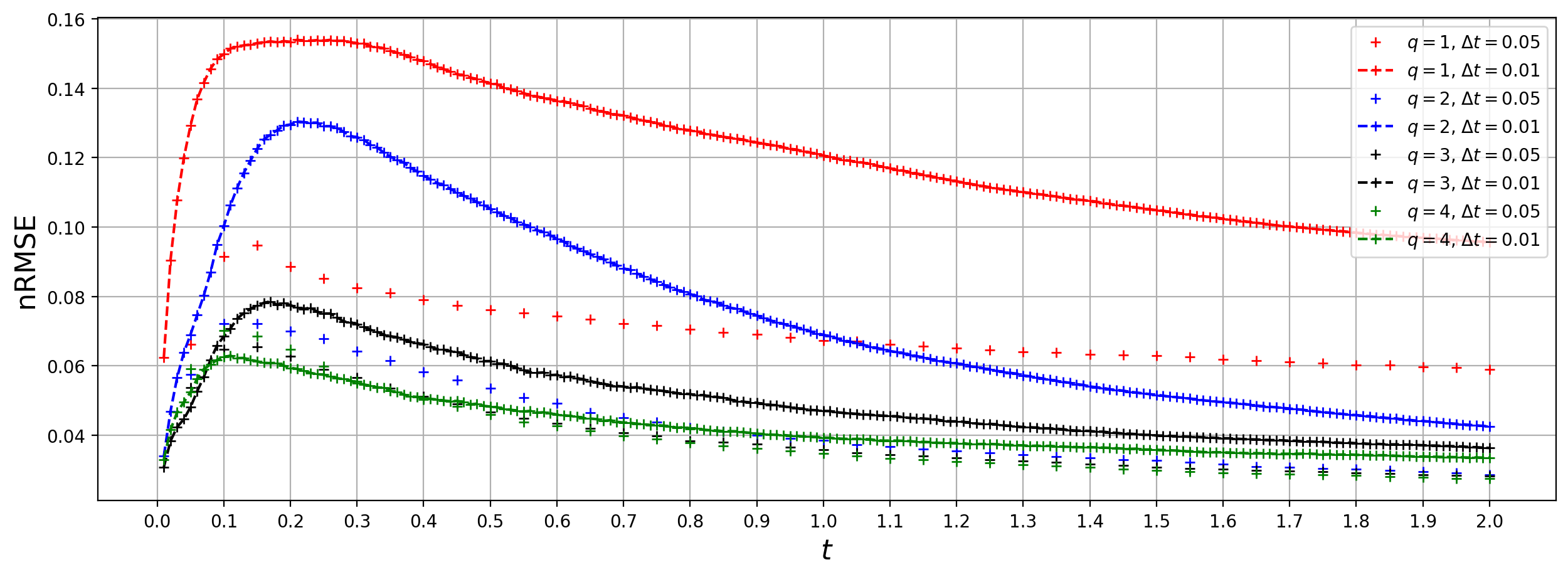}
  \caption{We show the nRMSE vs time when varying the stage $q$ of the RK algorithm to solve the ODE of Equation (\ref{eq:ODE}) on the Burgers' dataset with $\nu=0.001$. The same $q$ is used both at training and at inference. We see that by increasing $q$ not only we improve the predictions when using the same $\Delta t$ used during training ($\Delta t = 0.01)$ but we also get a better generalization in time.} 
  \label{fig:ODE_comparison_burger}
\end{figure}

\begin{figure}
  \centering
  \includegraphics[width=1.0\textwidth]{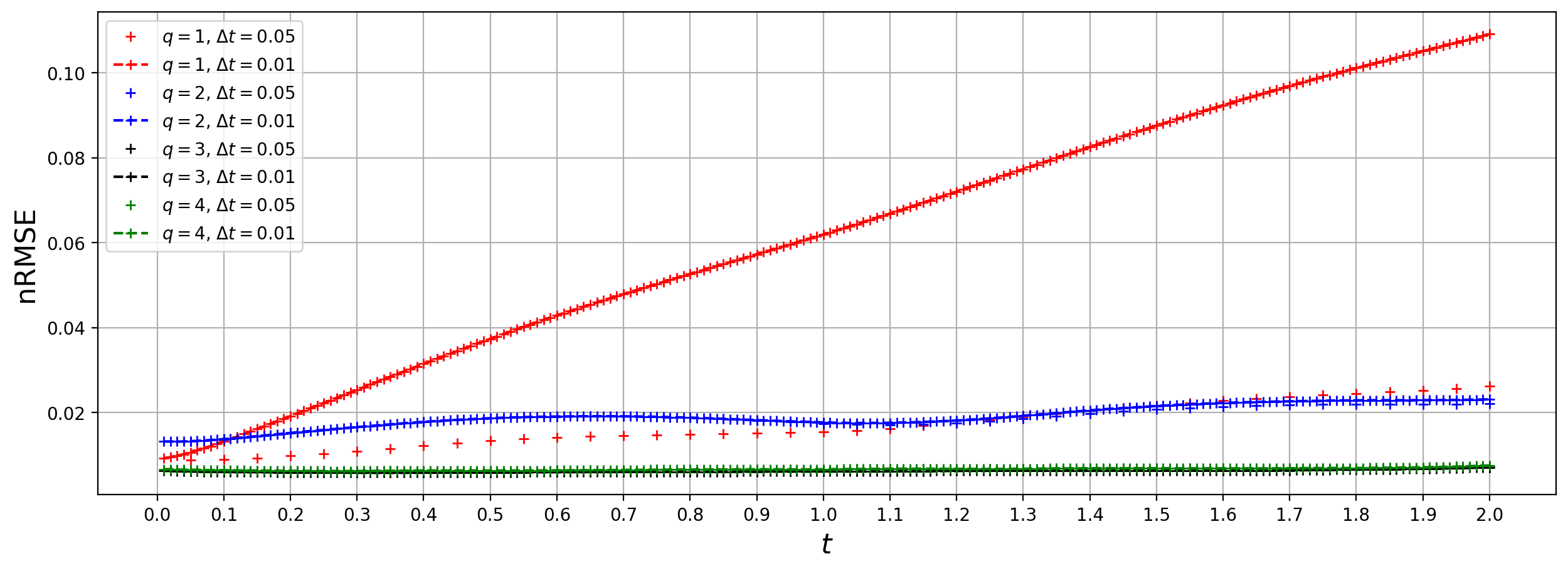}
  \caption{We show the nRMSE vs time when varying the stage $q$ of the RK algorithm to solve the ODE of Equation (\ref{eq:ODE}) on the Advection dataset with $\zeta=0.1$. The same $q$ is used both at training and at inference. We see that by increasing $q$ not only we improve the predictions when using the same $\Delta t$ used during training ($\Delta t = 0.01)$ but we also get a better generalization in time.} 
  \label{fig:ODE_comparison_advection}
\end{figure}

In Figure \ref{fig:ODE_comparison_burger} we show the effect of the stage $q$ of the RK algorithm used to solve Equation (\ref{eq:ODE}) for the Burgers' dataset with $\nu=0.001$. We see that by increasing $q$ not only the nRMSE is lowered, but also the gap between the trajectory of $\Delta t = 0.05$ (used during training)  and $\Delta t = 0.01$ is decreased, i.e., the larger the $q$ the better the generalization in time during inference. This is particularly clear when looking at the nRMSE over time of $q=3$ and $q=4$, since for $\Delta t =0.05$ they are almost the same, while for $\Delta t =0.01$ it is noticeably lower when $q=4$. In Figure \ref{fig:ODE_comparison_advection} we do the same experiment with the Advection dataset: here only for $q=1$ there is a big gap between the prediction at $\Delta t=0.05$ and $\Delta t=0.01$. 

\begin{figure}
  \centering
  \includegraphics[width=1.0\textwidth]{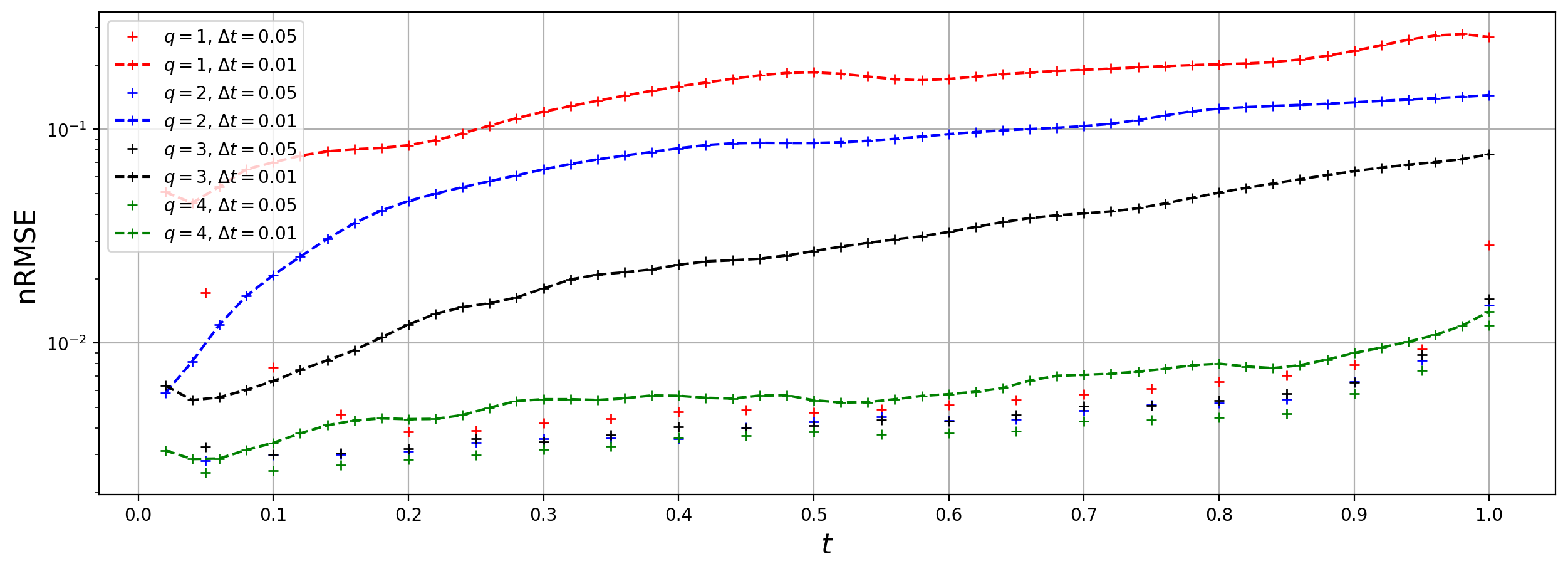}
  \caption{We show the nRMSE vs time when varying the stage $q$ of the RK algorithm to solve the ODE of Equation (\ref{eq:ODE}) on the Molenkamp dataset. The same $q$ is used both at training and at inference. We see that by increasing $q$ not only we improve the predictions when using the same $\Delta t$ used during training ($\Delta t = 0.01)$ but we also get a better generalization in time.} 
  \label{fig:ODE_comparison_molenkamp}
\end{figure}

\begin{figure}
  \centering
  \includegraphics[width=1.0\textwidth]{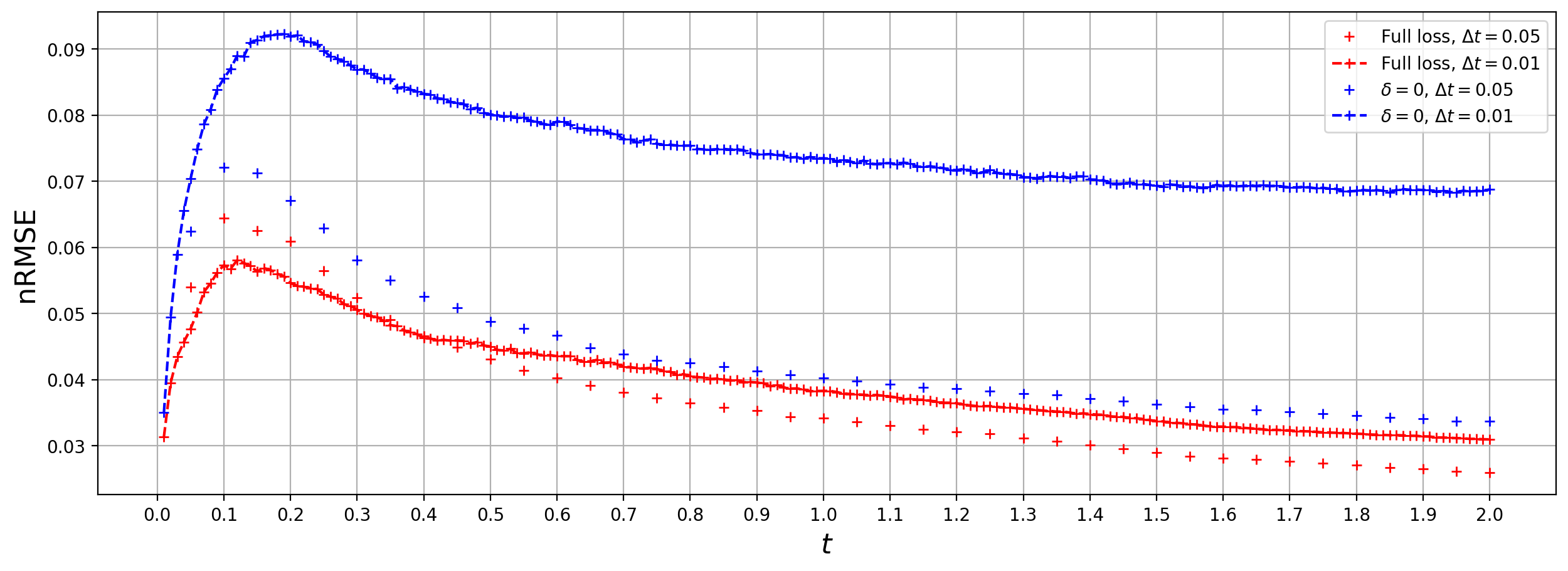}
  \caption{nRMSE over time when using $\mathcal{L}_3$ in the training (red curves) and when not (blue curves, $\delta=0$). The presence of $\mathcal{L}_3$ at training improves the generalization in time. $\Delta t = 0.05$ is the training time-step.} 
  \label{fig:no_l3_burger}
\end{figure}
In Figure \ref{fig:ODE_comparison_molenkamp} once again we show the same pattern for the Molenkamp dataset: increasing the value of $q$ results in a better cabability of the model to generalize in time during inference by taking a smaller $\Delta t$.

In Figure \ref{fig:no_l3_burger} we show a comparison of the nRMSE over time on the Burgers' dataset with $\nu=0.001$ between using the full loss $\mathcal{L}_{tr}$ and switching off $\mathcal{L}_3$ by setting $\delta=0$: while for $\Delta t =0.05$ (the one used at training) the two curves are comparable, for $\Delta t =0.05$ a huge gap is present. This result is in line with the reasoning that $\mathcal{L}_3$ helps the model to generalize in time, as explained in \ref{subsec:generalization_in_time}.
\section{Datasets}
\label{subsec:datasets}
Unless stated otherwise, the solutions of the PDEs described in this section used in the training come from \cite{takamoto2022pdebench}.
\subsection{1D Advection Equation}
\label{subsubsec:advection_dataset}
The 1D Advection Equation is a linear PDE which transports the initial condition with a constant velocity $\zeta$:
\begin{equation}
\label{eq:advection}
    \left\{
    \begin{aligned}
    &\partial_t s(\mathbf{x},t|\pmb{\mu})+\zeta \partial_x s(\mathbf{x},t|\pmb{\mu})= 0,\quad x\in(0,1),\, t\in(0,2]\\
    &  s(\mathbf{x},0|\pmb{\mu}) = s^0(\mathbf{x},\pmb{\mu}),\, x\in(0,1).
    \end{aligned}
    \right.
\end{equation}
Periodic boundary conditions are considered and as initial condition a super-position of sinusoidal waves is used:
\begin{equation}
    s^0(\mathbf{x},\pmb{\mu}) = \sum_{k_i=k_1,...,k_N}A_i \sin(k_i x+\phi_i),
\end{equation}
where $k_i = 2\pi\,{n_i}/L_x$ with ${n_1}$ being random integers, $L_x$ is the domain size, $A_i$ is a random number from the interval $[0,1]$ and $\phi_i$ is the phase chosen randomly in $(0,2\pi)$. We use 256 equidistant spatial points in the interval $[0,1]$ and for training 41 uniform timesteps in the interval $[0,2]$.
\subsection{1D Burgers' Equation}
The Burgers's equation is a non-linear PDE used in various modeling tasks such as fluid dynamics and traffic flows:
\begin{equation}
\label{eq:burger}
    \left\{
    \begin{aligned}
    &\partial_t s(\mathbf{x},t|\pmb{\mu})+\partial_x(s^2(\mathbf{x},t|\pmb{\mu})/2)-\nu/\pi\partial_{xx}s(\mathbf{x},t|\pmb{\mu})\quad x\in(0,1),\, t\in(0,2]\\
    &  s(\mathbf{x},0|\pmb{\mu}) = s^0(\mathbf{x},\pmb{\mu}),\, x\in(0,1),
    \end{aligned}
    \right.
\end{equation}
where $\nu$ is the diffusion coefficient. The initial conditions and the boundary conditions are the same as in Subsection \ref{subsubsec:advection_dataset}. We use 256 equidistant spatial points in the interval $[0,1]$ and for training 41 uniform timesteps in the interval $[0,2]$.
\subsection{2D Shallow Water Equations}
The 2D Shallow Water Equations are a system of hyperbolic PDEs derived from the Navier Stokes equations and describe the flow of fluids, primarily water, in situations where the horizontal dimensions (length and width) are much larger than the vertical dimension (depth):
\begin{equation}
\label{eq:sw}
    \begin{aligned}
    &\partial_t h+\partial_x h u + \partial_y h v = 0, \\
    &\partial_t h u +\partial_x \left( u^2 h + \frac{1}{2} g_r h^2 \right)+\partial_y u v h = -g_r h\partial_x b,  \\
    & \partial_t h v +\partial_x \left( v^2 h + \frac{1}{2} g_r h^2 \right)+\partial_y u v h = -g_r h\partial_y b,
    \end{aligned}
\end{equation}
where $u,v$ are the horizontal and vertical velocities, $h$ is the water depth and $b$ is a spatially varying bathymetry. $g_r$ is the gravitational acceleration. We use $128\times 128$ equidistant spatial points in the interval $[-1,1]\times[-1,1]$ and for training 21 uniform timesteps in the interval $[0,1]$, while the compared methods use 101 uniform timesteps in the interval $[0,1]$.
\subsection{2D Molenkamp test}
\label{subsec:molenkamp_test}
The Molenkamp test is a two dimensional advection problem, whose exact solution is given by a Gaussian function which is transported trough a circular path without modifying its shape. Here we add a reaction term which makes the Gaussian shape decay over time:
\begin{equation}
\label{eq:molenkamp}
    \begin{aligned}
    &\partial_t q(x,y,t)+u\partial_x q(x,y,t) + v\partial_y q(x,y,t)+\lambda_3 q(x,y,t) = 0 \\
    &q(x,y,0) = \lambda_1\,0.01^{\lambda_2 h(x,y,0)^2},\quad h(x,y,0)=\sqrt{(x-\lambda_4+\frac{1}{2})^2+(y-\lambda_5)^2},
    \end{aligned}
\end{equation}
with $u=-2\pi y$ and $v = 2\pi x$ and $(x,y)\in[-1,1]$. For this problem an exact solution exists:
\begin{equation}
    \begin{aligned}
        &q(x,y,t) = \lambda_1 0.01^{\lambda_2 h(x,y,t)^2}\exp^{-\lambda_3 t},\\
        &h(x,y,t) = \sqrt{(x-\lambda_4+\frac{1}{2}\cos(2\pi t))^2+(y-\lambda_5-\frac{1}{2}\sin(2\pi t))^2}.
\end{aligned}
\end{equation}
\begin{figure}[h!]
  \centering
  \includegraphics[width=1\textwidth]{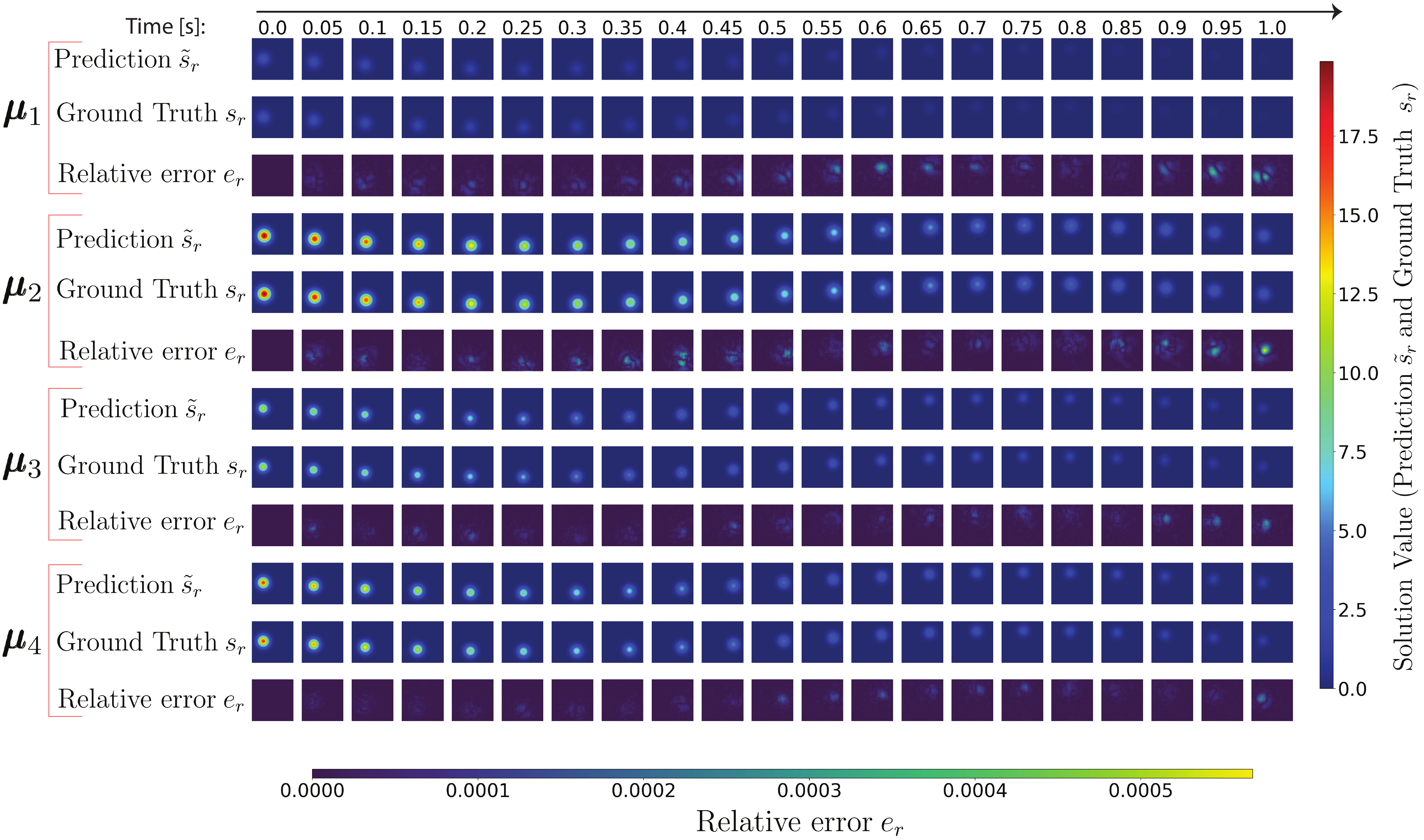}
  \caption{We show the predictions of our model over time on the Molenkamp test dataset for $4$ different combination of parameters $\pmb{\mu}_1$, $\pmb{\mu}_2$, $\pmb{\mu}_3$ and $\pmb{\mu}_4$. The vertical colorbar refers to the fields $s_r(\mathbf{x},t|\pmb{\mu})$ and $\tilde{s}_r(\mathbf{x},t|\pmb{\mu})$ (prediction and ground truth), while the horizontal one to the relative error $e_r$ of Equation \ref{eq:relative_error}.} 
  \label{fig:molenkamp_fields}
\end{figure}
\begin{figure}[h!]
  \centering
  \includegraphics[width=1\textwidth]{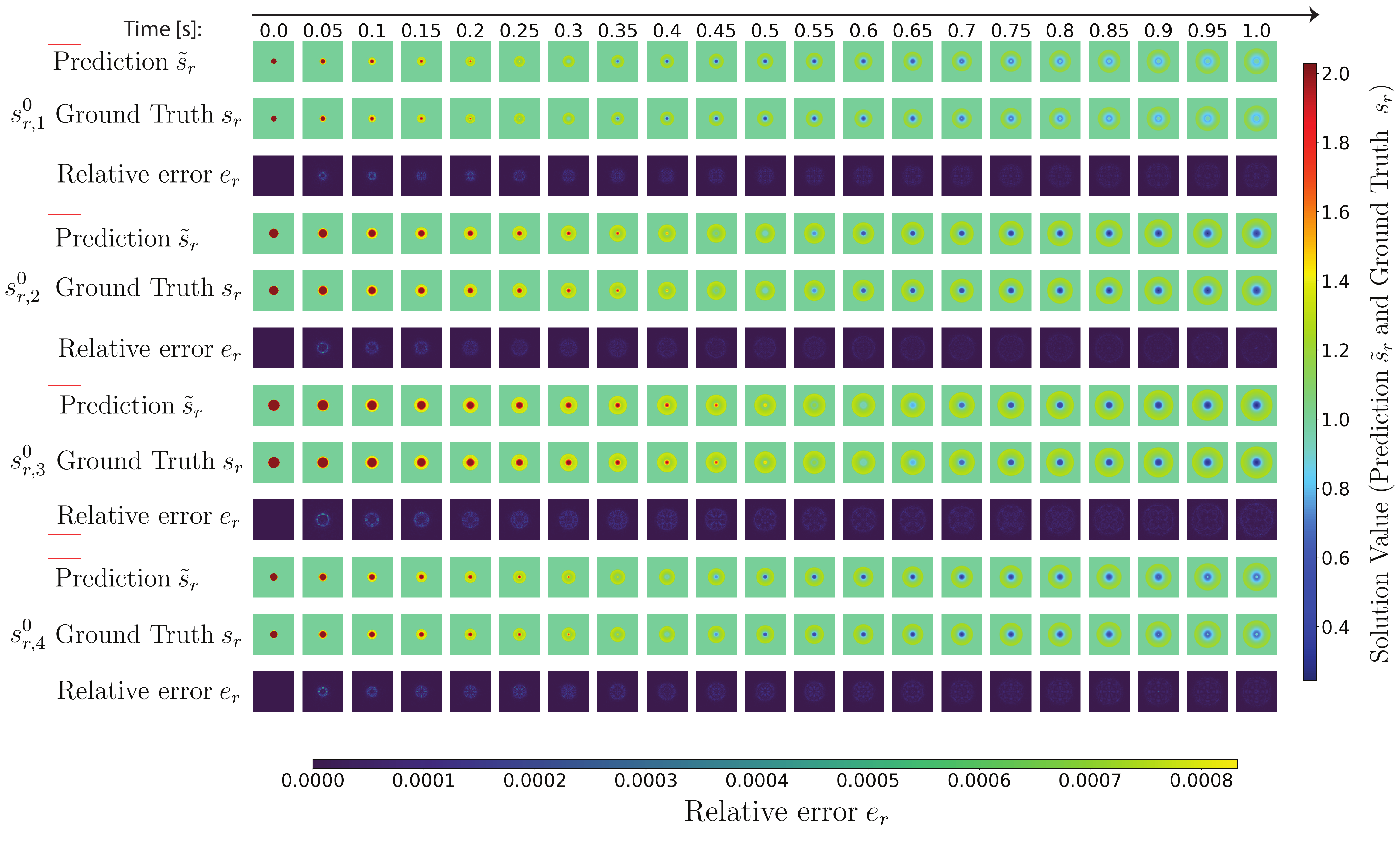}
  \caption{We show the predictions of our model over time on the Shallow-Water test dataset for $4$ different initial conditions $s^0_{r,1}$, $s^0_{r,2}$, $s^0_{r,3}$ and $s^0_{r,4}$. The vertical colorbar refers to the fields $\tilde{s}_r(\mathbf{x},t|\pmb{\mu})$ and $s_r(\mathbf{x},t|\pmb{\mu})$ (prediction and ground truth), while the horizontal one to the relative error $e_r$ of Equation \ref{eq:relative_error}.} 
  \label{fig:SW}
\end{figure}
The PDE depends on $5$ parameters $\lambda_1,...,\lambda_5$, which control the magnitude of the initial Gaussian, the size of the cloud, the spees of decay, the initial x coordinate and the initial y coordinate, respectively. In Table we show the parameters' ranges as it is done in \cite{Alsayyari_2021}: $\lambda_1\in[1,20]$, $\lambda_2\in[2,4]$, $\lambda_3\in[1,5]$, $\lambda_4\in[-0.1,0.1]$, $\lambda_5\in[-0.1,0.1]$. We use $128\times 128$ equidistant spatial points in the interval $[-1,1]\times[-1,1]$ and for training 21 uniform timesteps in the interval $[0,1]$.
\subsection{Metrics}
 We use as testing metric the Normalized-Mean-Squared-Root-Error (nRMSE), defined as
\begin{equation}
\label{eq:nRMSE}
    \text{nRMSE} =\frac{1}{N_u\,N_{\pmb{\mu}}F}\,\sum_{i=1}^{N_u}\sum_{p=1}^{N_{\pmb{\mu}}}\sum_{j=1}^F \frac{||s_r(\mathbf{x},t_j|\pmb{\mu}_p,s_{r,i}^0)-\tilde{s}_r(\mathbf{x},t_j|\pmb{\mu}_p,s_{r,i}^0))||_2}{||s_r(\mathbf{x},t_j|\pmb{\mu}_p,s_{r,i}^0)||_2},
\end{equation}
where $N_u$, $N_{\pmb{\mu}}$ and $F$ are the number of initial conditions, parameter instances and time steps used at testing, respectively; $s^0_{r,i}$ stands for the $i$th initial condition used at testing. We also define the \textit{relative error} $e_r$, i.e., a more spatially descriptive error measure between the prediction of a field $\tilde{s}_r(\mathbf{x},t|\pmb{\mu})$ and the expected field $s_r(\mathbf{x},t|\pmb{\mu})$:
\begin{equation}
\label{eq:relative_error}
    e_r = \frac{|s_r(\mathbf{x},t|\pmb{\mu})-\tilde{s}_r(\mathbf{x},t|\pmb{\mu})|}{||s_r(\mathbf{x},t|\pmb{\mu})||_2},
\end{equation}
where the numerator is the point-wise absolute value of the difference between $\tilde{s}_r(\mathbf{x},t|\pmb{\mu})$ and $s_r(\mathbf{x},t|\pmb{\mu})$ (hence it has the same dimensionality as $s_r(\mathbf{x},t|\pmb{\mu})$), while the denominator is a scalar.
\section{Additional images}

\begin{figure}
  \centering
  \includegraphics[width=0.9\textwidth]{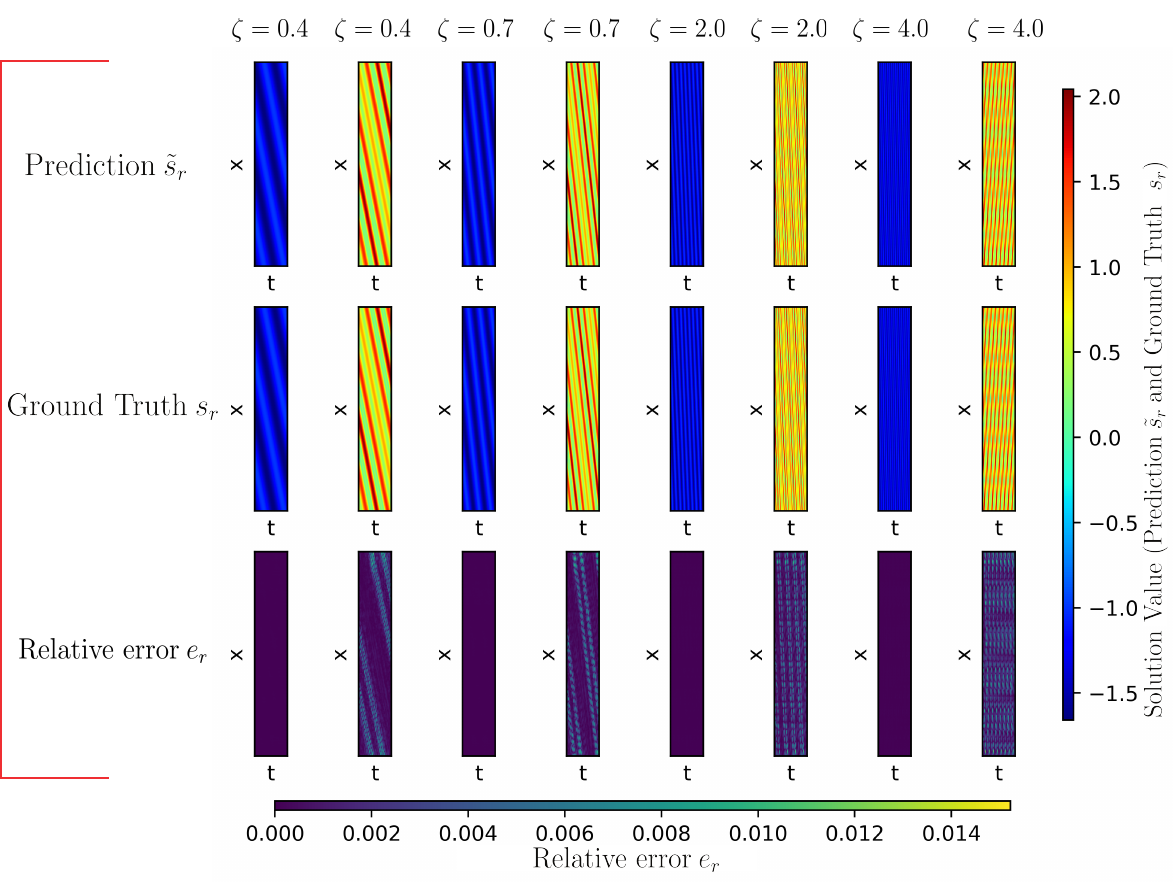}
  \caption{We show the predictions of our model on the parametric Advection dataset for $2$ different initial conditions and $4$ different velocities: $\zeta=0.4$, $\zeta=0.7$, $\zeta=2.0$ and $\zeta=4.0$. Each plot is a heat map with time $t$ on the horizontal axis and space $x$ on the vertical axis. The vertical colorbar refers to the fields $\tilde{s}_r(\mathbf{x},t|\pmb{\mu})$ and $s_r(\mathbf{x},t|\pmb{\mu})$ (prediction and ground truth), while the horizontal one refers to the relative error $e_r$ of Equation \ref{eq:relative_error}.} 
  \label{fig:advection_x_t}
\end{figure}
In Figure \ref{fig:molenkamp_fields} we show the predictions of our model on the Molenkamp test for $4$ different parameter values: $\pmb{\mu}_1= [2.4522 ,2.3731 ,  2.7912  ,0.0533,0.01250]$, $\pmb{\mu}_2=[19.8578  ,   2.5791  , 1.9388 ,  0.0959, -0.0857]$, $\pmb{\mu}_3=[11.7423 ,   3.9285   , 2.5638 ,  0.0384,  0.0200]$ and $\pmb{\mu}_4=[16.8555  , 3.4449  , 2.6506  , 0.0502 , 0.0423]$.
In Figure \ref{fig:SW} we show the predictions of our model on the Shallow-Water test case for $4$ different initial conditions $s^0_{r,1}$, $s^0_{r,2}$, $s^0_{r,3}$ and $s^0_{r,4}$. 
\begin{figure}
  \centering
  \includegraphics[width=1.0\textwidth]{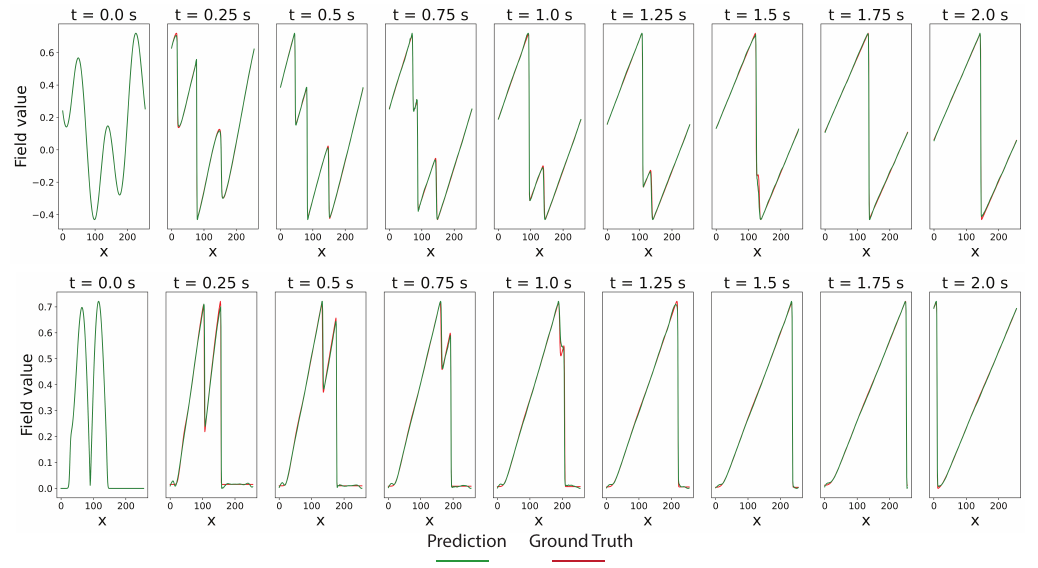}
  \caption{We show the predictions of our model on the Burgers' dataset when $\nu=0.001$. The two rows correspond to two different initial conditions.} 
  \label{fig:burgers_x_t}
\end{figure}
In Figure \ref{fig:advection_x_t} we show the predictions of our model on the 1D Advection test case for $2$ different initial conditions and  $4$ different velocities: $\zeta=0.4$, $\zeta=0.7$, $\zeta=2.0$ and $\zeta=4.0$. Finally, in Figure \ref{fig:burgers_x_t} we show the prediction for the Burgers case when $\nu=0.001$ for two different initial conditions.

\end{document}